%% file: papermain.tex
\documentclass[10pt,twocolumn,letterpaper]{article}

\usepackage[pagenumbers]{cvpr}
\usepackage{times}
\usepackage{epsfig}
\usepackage{graphicx}
\usepackage{amsmath}
\usepackage{amssymb}
\usepackage{booktabs}
\usepackage{caption}

\usepackage{graphicx}
\usepackage{booktabs}
\usepackage{array}
\usepackage{xcolor}
\usepackage{scalerel}
\usepackage{enumitem}
\usepackage{xspace}
\usepackage[font={small},labelfont=bf]{caption}
\usepackage{tabu}

\newcommand\jtp[1]{{\color{green}}}

\usepackage{multirow}

\usepackage{hyperref}

\makeatletter
\DeclareRobustCommand\onedot{\futurelet\@let@token\@onedot}
\def\@onedot{\ifx\@let@token.\else.\null\fi\xspace}

\def\eg{\emph{e.g}\onedot}

\def\etal{\emph{et al}\onedot}
\makeatother

\begin{document}

\newcommand{\papertitle}{VPFusion}
\title{\papertitle: Joint 3D Volume and Pixel-Aligned Feature Fusion for Single and Multi-view 3D Reconstruction}

\author{Jisan Mahmud \qquad Jan-Michael Frahm \\
University of North Carolina at Chapel Hill \\
{\tt\small \{jisan,jmf\}@cs.unc.edu}}

\twocolumn[{%
\renewcommand\twocolumn[1][]{#1}%
\maketitle
\begin{center}
\centering
\vspace{-8mm}
\includegraphics[width=0.88\linewidth]{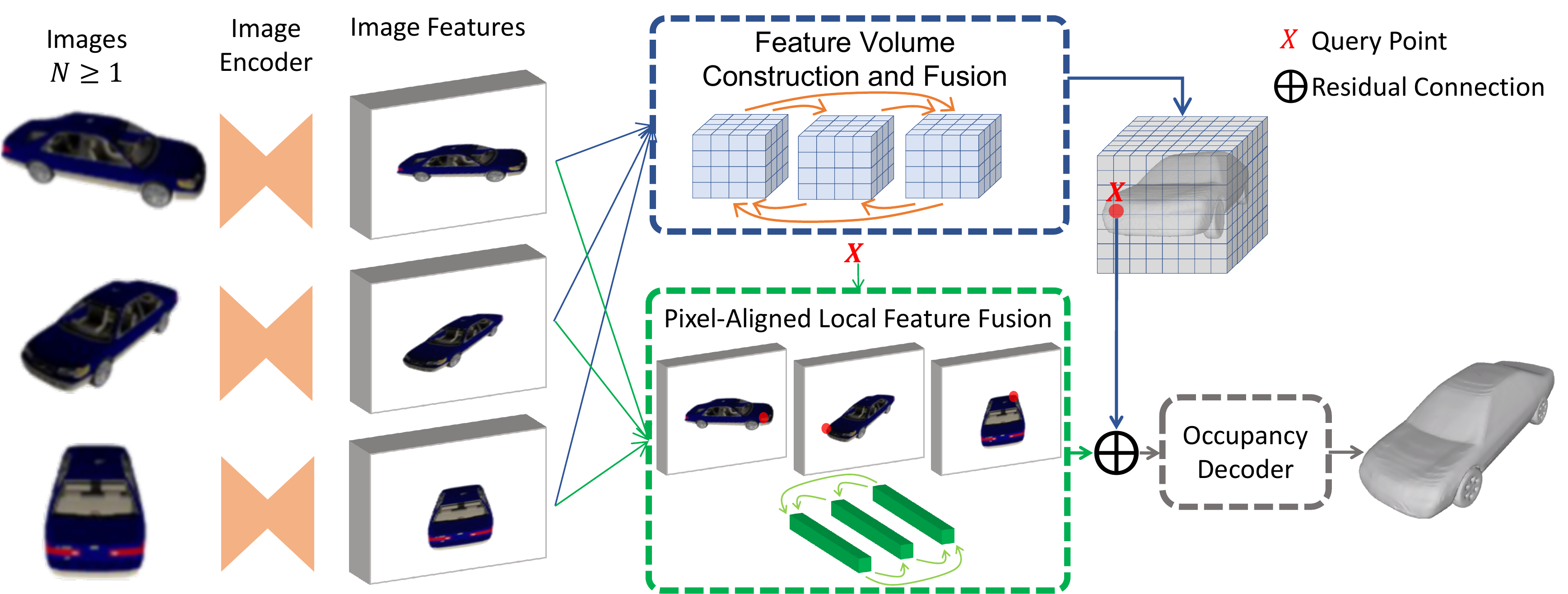}
\vspace{-2mm}
\captionof{figure}{
\label{fig:mainfig}
Overview of \papertitle~3D reconstruction from $N \geq 1$ images.
The feature volume construction and fusion module takes the image features, constructs 3D feature volume for each and fuses them across different views.
This provides a 3D-structure-aware and multi-view aware context.
The pixel-aligned local feature fusion fuses pixel-aligned features for given 3D point across views to provide fine local detail.
Finally, these features are residually added for the 3D point and decoded to an occupancy value.
\vspace{-1mm}
}
\label{fig:ourmainmodel}
\end{center}%
}]

\thispagestyle{empty}

\begin{abstract}
\vspace{-2mm}
We introduce a unified single and multi-view neural implicit 3D reconstruction framework \papertitle.
\papertitle~attains high-quality reconstruction using both - 3D feature volume to capture 3D-structure-aware context, and pixel-aligned image features to capture fine local detail.
Existing approaches use RNN, feature pooling, or attention computed independently in each view for multi-view fusion.
RNNs suffer from long-term memory loss and permutation variance, while feature pooling or independently computed attention leads to representation in each view being unaware of other views before the final pooling step.
In contrast, we show improved multi-view feature fusion by establishing transformer-based pairwise view association.
In particular, we propose a novel interleaved 3D reasoning and pairwise view association architecture for feature volume fusion across different views.
Using this structure-aware and multi-view-aware feature volume, we show improved 3D reconstruction performance compared to existing methods.
\papertitle~improves the reconstruction quality further by also incorporating pixel-aligned local image features to capture fine detail.
We verify the effectiveness of \papertitle~on the ShapeNet and ModelNet datasets, where we outperform or perform on-par the state-of-the-art single and multi-view 3D shape reconstruction methods.

\end{abstract}

\input{introduction}
\input{relatedworks}
\input{methodology}
\input{experiments}

\input{conclusion}

{\small
\bibliographystyle{ieee_fullname}
\bibliography{bibliography_jisan}
}

\end{document}

\newcommand{\papertitle}{VPFusion}
\title{Supplementary - \papertitle: Joint 3D Volume and Pixel-Aligned Feature Fusion for Single and Multi-view 3D Reconstruction}

\author{Jisan Mahmud \qquad Jan-Michael Frahm \\
University of North Carolina at Chapel Hill \\
{\tt\small \{jisan,jmf\}@cs.unc.edu}}

\maketitle

\begin{figure*}[h!]
  \centering
  \includegraphics[width=0.7\linewidth]{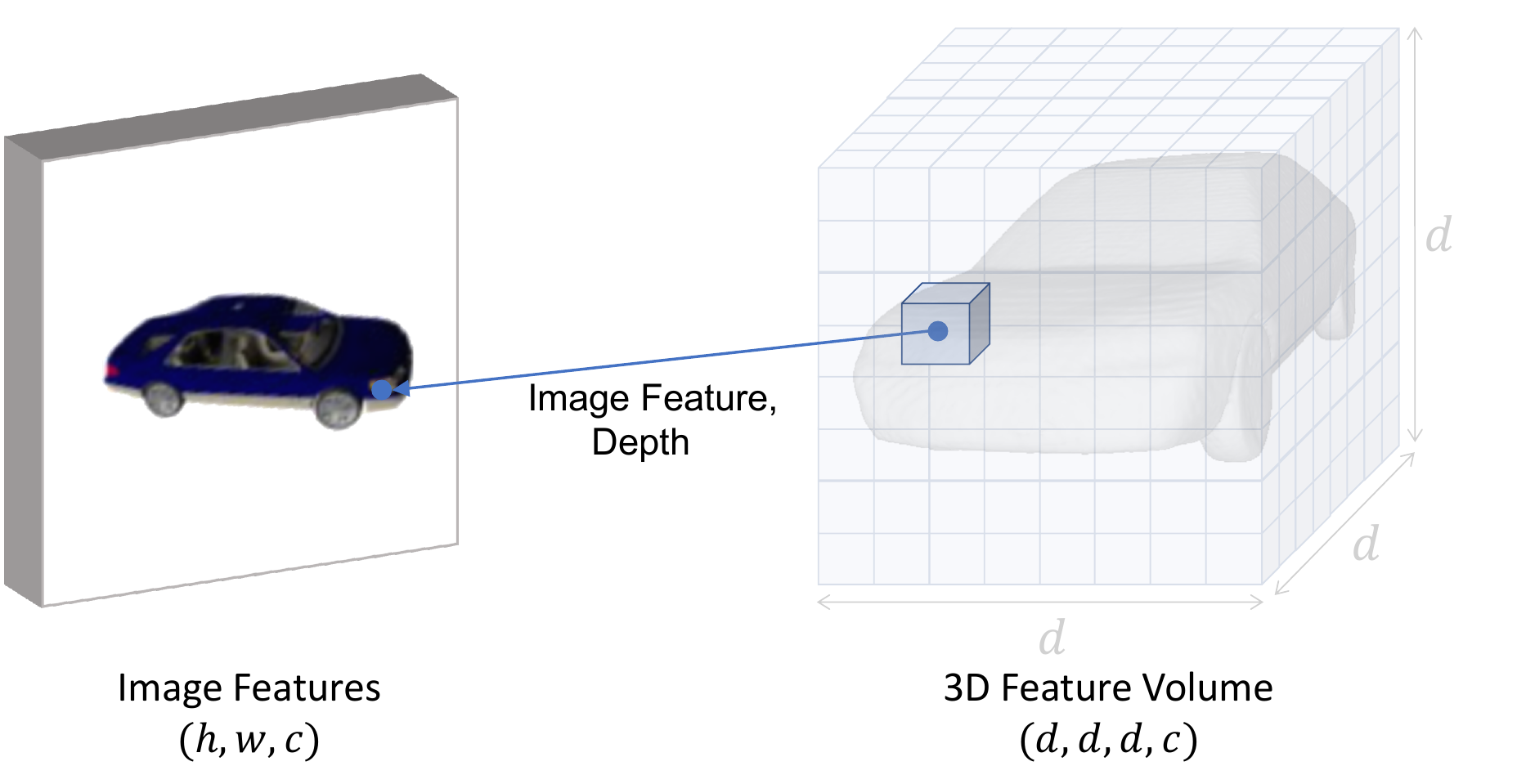}
  \caption{
  \label{fig:featvolrec}
  3D feature volume construction.
  }
\end{figure*}

\input{supplementary_materials/implicit_choice}

\section{Implementation Details}

\subsection*{Constructing 3D Feature Volume}
\papertitle~creates a 3D feature volume for each image $I^{(i)}$ (Fig. \ref{fig:featvolrec}) as denoted in Sec. 3.1.1 of the main paper.
First, we uniformly divide the 3D volume in a $d\times d \times d$ grid.
We sample feature for each grid cell by projecting the cell center onto feature map $g^{(i)}$ and sampling bi-linearly.
Since multiple 3D points can project to the same feature map location, we also extract the depth of these cell centers to incorporate spatial distinction.
Instead of using the depth values directly, we instead utilize the normalized and positionally encoded depth values.
The normalization is a simple fixed scaling operation.
Given normalized depth $d$, the positional encoding is defined as:

\begin{align}
\begin{split}
    \gamma(d) = & \displaystyle [ \sin{\left(2^0 \pi d\right)}; \cos{\left(2^0 \pi d\right)}; \cdots  \\ & \qquad ;\sin{\left(2^{L - 1} \pi d\right)}; \cos{\left(2^{L - 1} \pi d\right)} \displaystyle ]
\end{split}
\end{align}

Here, $L$ refers to the number of frequency bands used in the positional encoding.
This encoding mechanism is similar to \cite{mildenhall2020nerf}.
While applying standard MLP to encode positional values fails to learn high-frequency information (as studied in \cite{tancik2020fourier}), this type of positional encoding can circumvent the problem by directly encoding high-frequency components as input signals.
In all experiments, we use $L=11$ frequency bands for the positional encoder.
Each grid cell features are then concatenated with the corresponding positional encoded depth features; followed by a small MLP.
This operation is performed for each grid cell, and this constructs the 3D feature volume $G^{(i)} \in \mathbb{R}^{d \times d \times d \times c}$.
$c$ is the feature length dimension for each grid cell.

\begin{figure*}[ht!]
  \centering
  \includegraphics[width=0.8\linewidth]{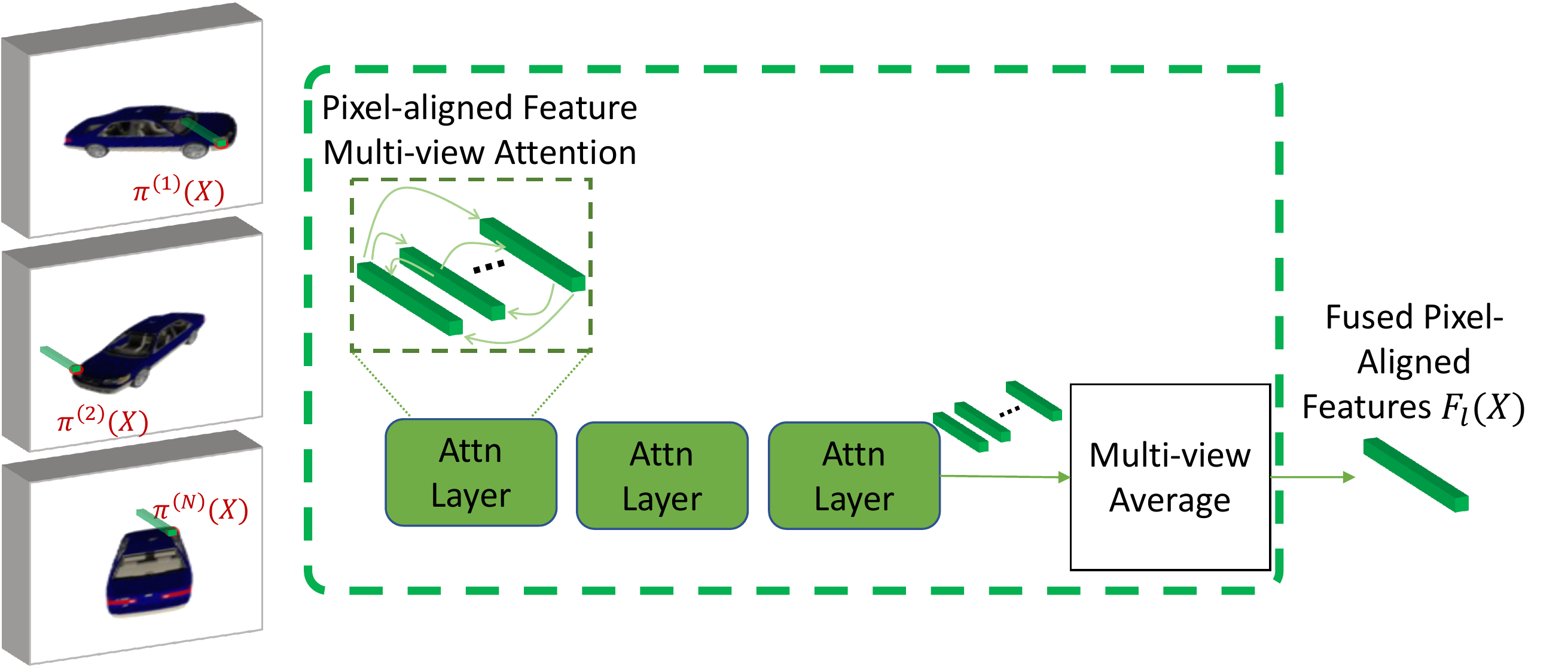}
  \caption{
  \label{fig:pixelalignedfusionfig} Multi-view pixel-aligned local feature fusion to extract fine local detail across all images.
Given some query 3D point $X$, this module projects $X$ onto different image feature maps and extracts pixel aligned features.
Next, it applies multi-view attention using some transformer self-attention layers, followed by multi-view averaging.
Here, $\pi^{(i)}$ represents the 2D projection operation onto image $I^{(i)}$}.
\end{figure*}

\begin{figure*}[ht!]
  \centering
  \includegraphics[width=0.745\linewidth]{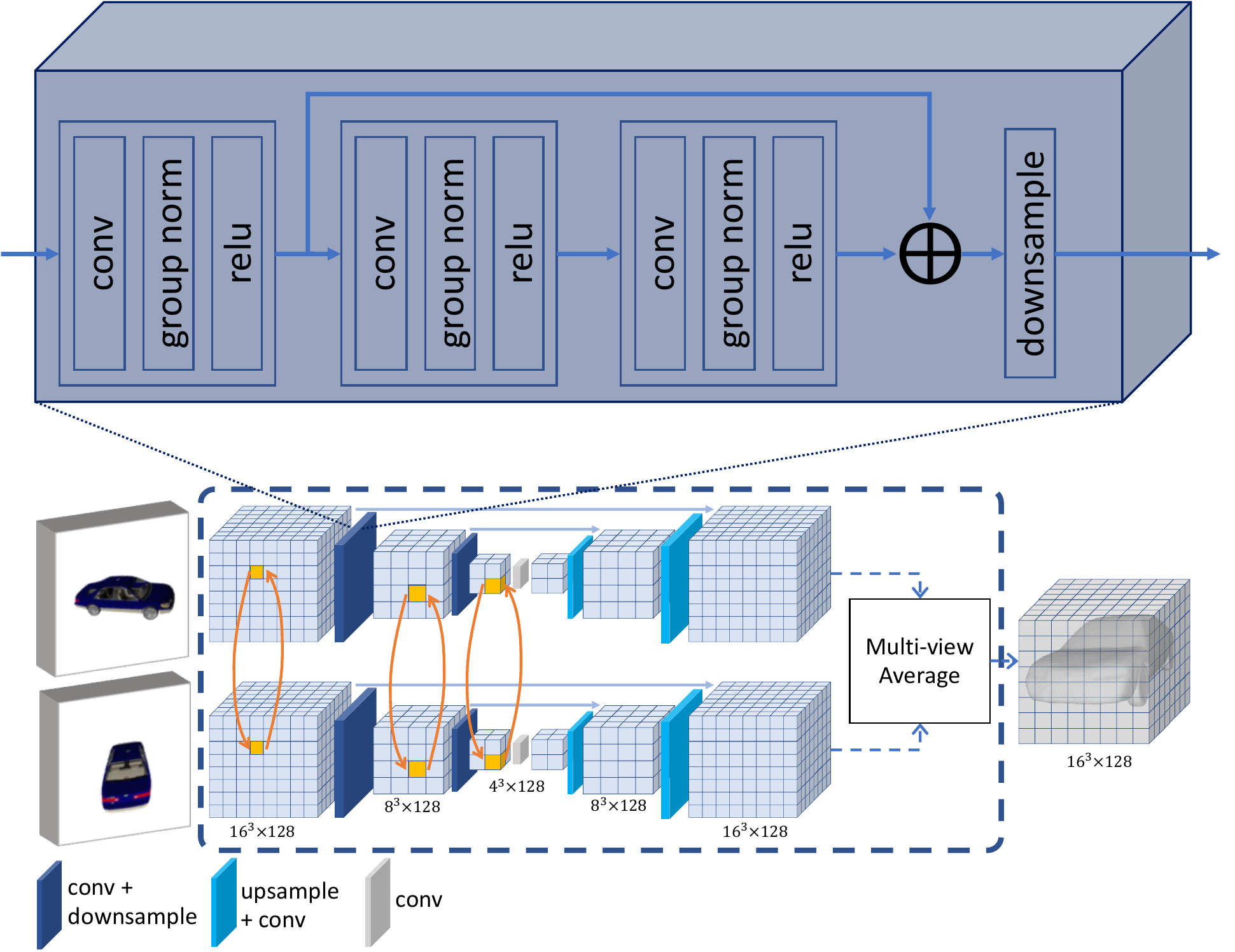}
  \caption{
  \label{fig:unetconvblock} Overview of a U-Net encoder block of our interleaved 3D-UNet and multi-view feature association network for \papertitle$_D$.
  All group normalization \cite{wu2018group} layers use 8 groups.
  The residual connection is followed by a downsampling layer, using a 3D convolution with $(2 \times 2 \times 2)$ kernel and $(2 \times 2 \times 2)$ stride.
  }
\end{figure*}
\subsection*{Feature Extraction from 3D Feature Volume}
After the feature volume construction and multi-view reasoning, \papertitle~utilizes the multi-view fused multi-view 3D feature volume features $F_g$ to provide 3D-structure-aware and multi-view aware features (main paper Sec. 3.1).
Given a 3D point $X$, we tri-linearly sample $F_g$ to obtain the corresponding feature $F_g(X)$.
An alternative possibility is performing the tri-linear interpolation after decoding.
In this case, the decoder would input the latent codes of 3D feature volume locations closest to $X$ and the relative location of $X$ w.r.t. the feature volume location, and predict the occupancy of $X$ in each such case.
Then these predicted occupancy values can be tri-linearly interpolated to compute the final occupancy of $X$.
However, this is inefficient and requires upto 8 occupancy decoding for a given 3D point. 
Another alternative would be to instead create a grid of local shape features similar to DeepLS \cite{chabra2020deep}.
DeepLS uses a set of local shape codes, where each shape code encodes the geometry of a distinct local shape.
To ensure smooth transition between adjacent local shapes, it overlaps the local shapes to their neighbors to some extent.
While more efficient than the former, in this fashion one can still require decoding upto 6 occupancy values for a given point $X$.
We instead choose to perform tri-linear interpolation on 3D feature volume $F_g$ directly (similar to \cite{peng2020convolutional}).
This requires decoding only one feature for the given 3D point, and also ensures continuous transition between different locations of the 3D feature volume.
This is also attested by the good reconstruction performances of \papertitle$_D$ (main paper Sec. 4).

\subsection*{Network Architecture}
Fig.~\ref{fig:pixelalignedfusionfig} illustrates the pixel-aligned local features fusion as described in Sec. 3.2 of the main paper.
The module consists of 3 transformer self-attention layers, where we use \texttt{n\_head=8} and \texttt{feed\_forward\_dim=128}.

Fig.~\ref{fig:unetconvblock} shows a U-Net encoder block of our proposed interleaved 3D-UNet and multi-view feature association network (described in Sec. 3.1.2 of the main paper).
The U-Net decoder blocks are designed similarly.
The only difference is - instead of downsampling, we apply upsampling at the end of the decoder block using a fractionally strided 3D convolution \cite{long2015fully} with $(2 \times 2 \times 2)$ kernel and $(2 \times 2 \times 2)$ stride.
At the middle layer of the U-Net (the layer colored gray) we apply a similar block without any downsampling or upsampling at its end.

For any sequence transformer in our work (where we incorporate the positional encoding), we compute sinusoidal positional encoding with 7 bands with exponential band separation of 2.
In all such cases, we concatenate positional information to the corresponding feature, followed by a small MLP to retain the dimensionality of the original feature.

For positional encoding of a 3D point or its depth, we always use a sinusoidal positional encoding with 11 bands with exponential band separation of 2.

\subsubsection*{\papertitle$_A$, \papertitle$_B$, \papertitle$_C$}

\begin{figure*}[h!]
\begin{subfigure}{1.0\linewidth}
\centering
\includegraphics[width=1.0\linewidth]{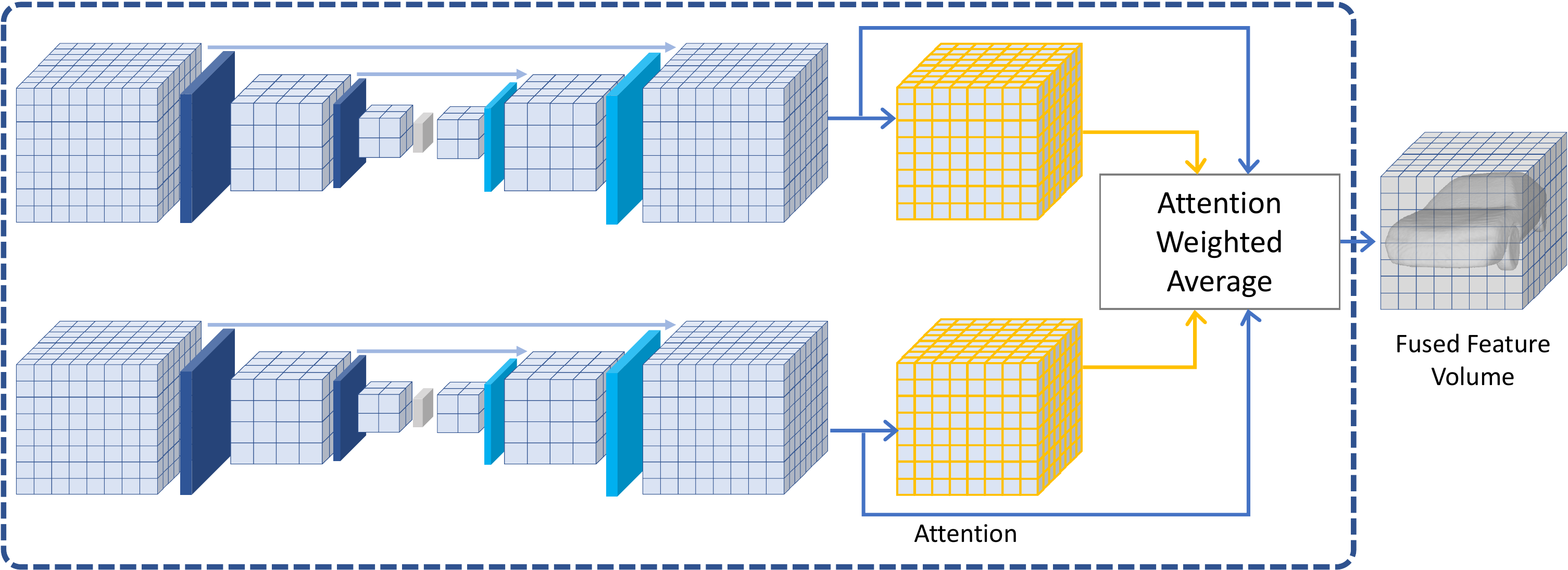}
\caption{\papertitle$_A$}
\end{subfigure}\\[1ex]
\begin{subfigure}{1.0\linewidth}
\centering
\includegraphics[width=0.85\linewidth]{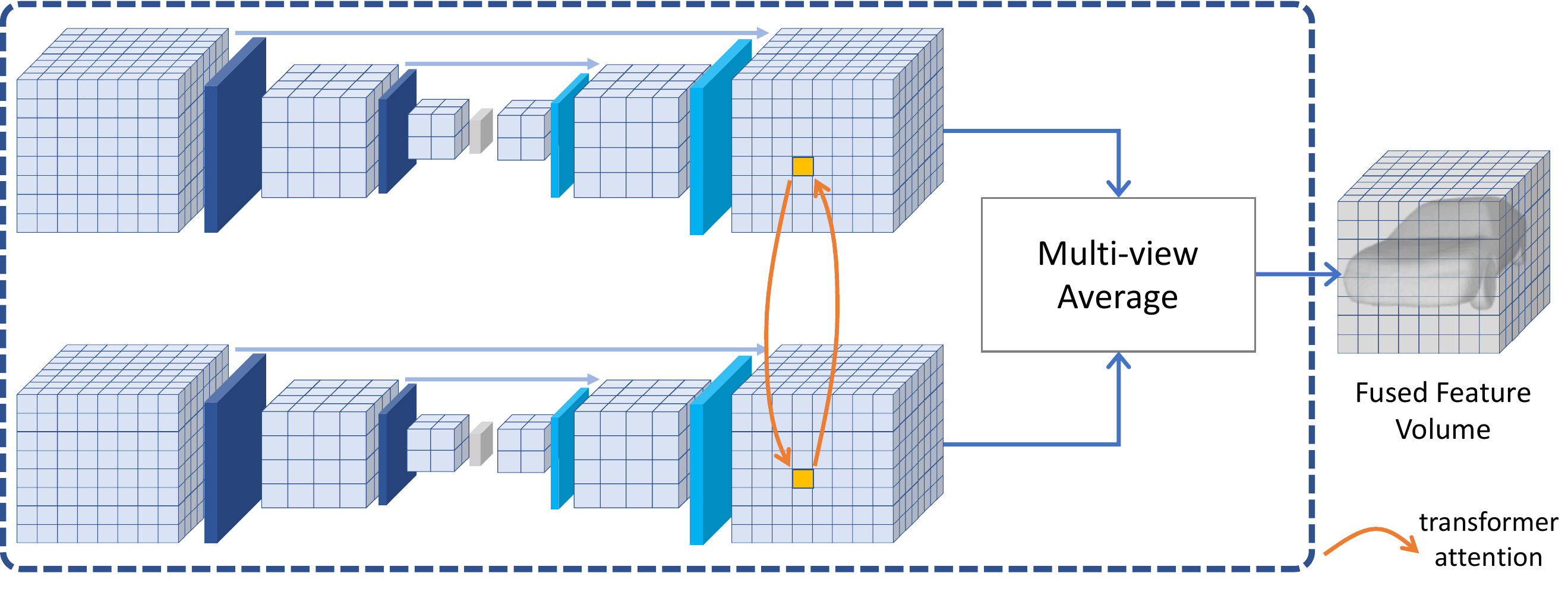}
\caption{\papertitle$_B$}
\end{subfigure}\\[1ex]
\begin{subfigure}{\linewidth}
\centering
\includegraphics[width=1.0\linewidth]{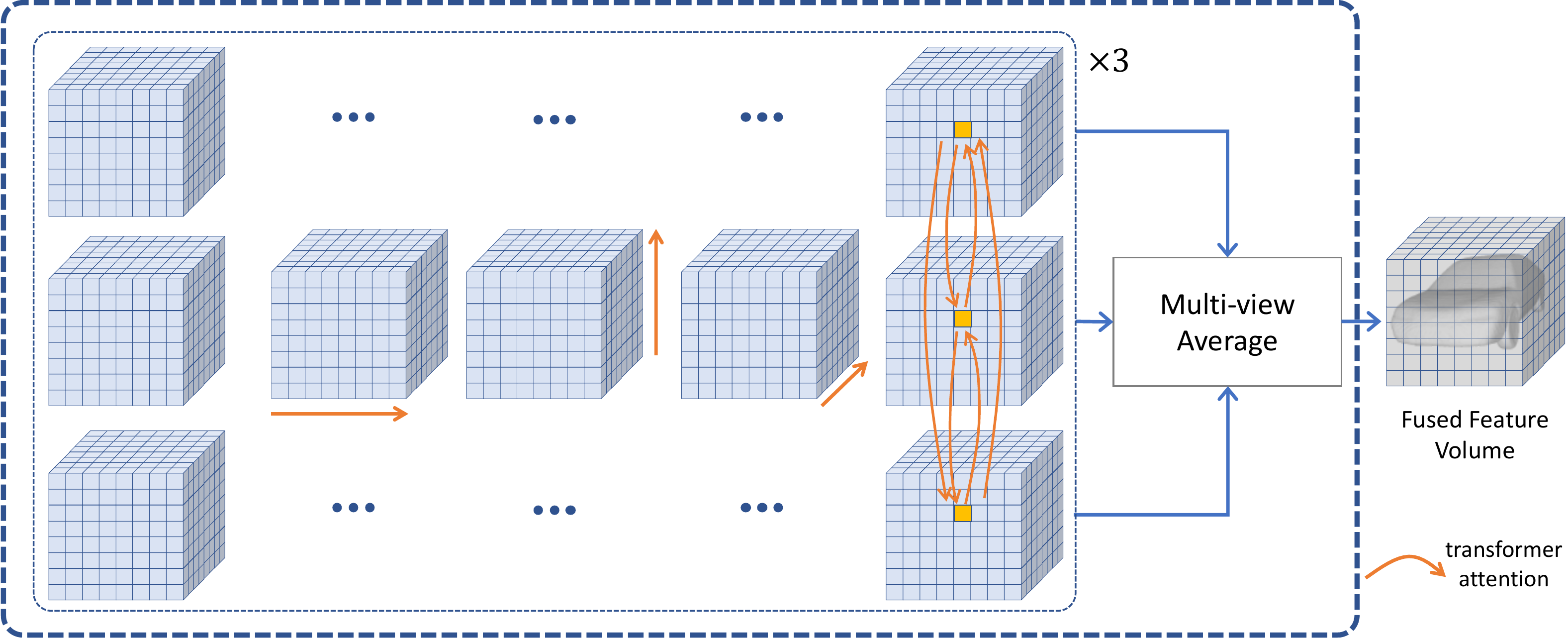}
\caption{\papertitle$_C$}
\end{subfigure}
\caption{Illustration of 3D-structure and multi-view reasoning of \papertitle$_{A-C}$.}
\label{fig:vpfusionalt}
\end{figure*}

For 3D-structure and multi-view reasoning, we have experimented with several possible network architectures (main paper Sec. 3.1.2).
Here, we illustrate \papertitle$_{A-C}$ in Fig. \ref{fig:vpfusionalt}.
\papertitle$_A$ performs 3D reasoning across all views using 3D U-Net, followed by independently computed attentions score.
Weighted attentions are used to aggregate multi-view information in this architecture.
\papertitle$_B$ similarly performs 3D reasoning across all views first, but then applies transformer-based attention across multiple views to aggregate multi-view information.
This allows the representation from each view to `observe' one another before fusion.
\papertitle$_C$ on the other hand, proposes to interleave the 3D-structure and multi-view reasoning by applying axial transformer attention, which we demonstrate to lead to higher quality reconstruction compared to \papertitle$_A$ and \papertitle$_B$.
We however obtain the highest quality 3D reconstruction using \papertitle$_D$ and \papertitle$_{Full}$ by applying interleaved 3D-structure and multi-view reasoning at different scales, as discussed in Sec. 3, 4 of the main paper.

\subsubsection*{Preparing ModelNet}
For the ModelNet40 \cite{wu20153d,SZB17a} dataset, the objects are scaled to fit into a unit cube centered at origin.
To render out the objects, we place a directional light source 1.2 units away from origin, at $30^{\circ}$ elevation, $90^{\circ}$ azimuth and directed at the origin.
24 renders are created for each object with camera placed 1.2 units away from the origin, with random azimuth and with elevation between $\left[ 15^{\circ}, 60^{\circ} \right]$.
We render out images with $224 \times 224$ resolution similar to the ShapeNet dataset of 3D-R2N2 \cite{choy20163d}.
\subsection*{Training}
For each object during the training, we sample 2048 3D points in a cube centered at the origin and with side length of 1.1.
The sampling strategy is denoted in Sec.~4.1 of the main paper.
The sampled points are a mixture of random uniform points in the bounded volume, and close-to-object surface points with a 1:5 ratio.
For close-to-object surface points, we sample points $\sim \mathcal{N}(0,\,\sigma^{2})\,$ away from random surface points, where $\sigma = 0.03$.

We use the AdamW \cite{loshchilov2018decoupled} optimizer with varying learning rate and weight decay of 0.0001.
We use a linear learning rate warmp, where from iteration 1 until 10K, the learning rate increases linearly from 0.0001 to 0.001.
From iteration 10K to 100K, the learning rate decreases linearly from 0.001 to 0.0001.
Then using learning rate of 0.0001, the network is trained until convergence (roughly 1 million iterations for both datasets).
We perform flipping images along the horizontal axis as the only form of data augmentation.

\input{tables/timingtable}

\section{Time Complexity}
\papertitle$_D$ attains significantly faster performance than PiFU \cite{saito2019pifu} (Tab. \ref{tab:timingtable} ), while achieving superior reconstruction quality especially from small number of observations (Tab. 2, 3 of the main paper).
PiFU performs relatively intensive feature manipulation for any 3D point across all views. 
On the other hand, \papertitle$_D$ applies attention only on the coarse 3D volumes across different views, while the decoder utilizes a small network.
As a result, it obtains significantly faster inference time.
\papertitle$_D$ has slightly higher inference time complexity compared to ONet \cite{mescheder2019occupancy}, but obtains much better reconstruction quality (\eg~0.664 vs 0.571 mIoU on ShapeNet). 
\papertitle$_{Full}$ further improves the reconstruction, but requires more time.
\papertitle$_{Full}$ has comparable time complexity as PiFU, while achieving better reconstruction quality from any number of views.

\input{tables/singleviewpix3d}

\section{Additional Quantitative Results}
Tab.~\ref{singleviewreconstructionpix3d} shows the synthetic to real world adaptation performance of our method (\papertitle$_D$) compared to some of the prior methods.
Here, all methods were all trained on ShapeNet, and evaluated on Pix3D.
We followed the protocol of \cite{mescheder2019occupancy}, where we mask out the background and crop the image to keep the object roughly in the middle.
Only 4 object classes (chair, desk, sofa, table) oxf Pix3D are used since these classes are similar to the ones in ShapeNet.
PiFU \cite{saito2019pifu} did not perform well in the adaptation test (mIoU of 0.118).
Presumably, the lack of 3D-structure understanding and sole reliance on pixel-level features prevents this method from good adaptation on real data.
Tab.~\ref{singleviewreconstructionpix3d} shows that compared to other methods, our method demonstrates improved adaptation performance in most cases.

\section{Additional Qualitative Results}
\begin{figure*}[!h]
\centering
\includegraphics[width=0.96\linewidth]{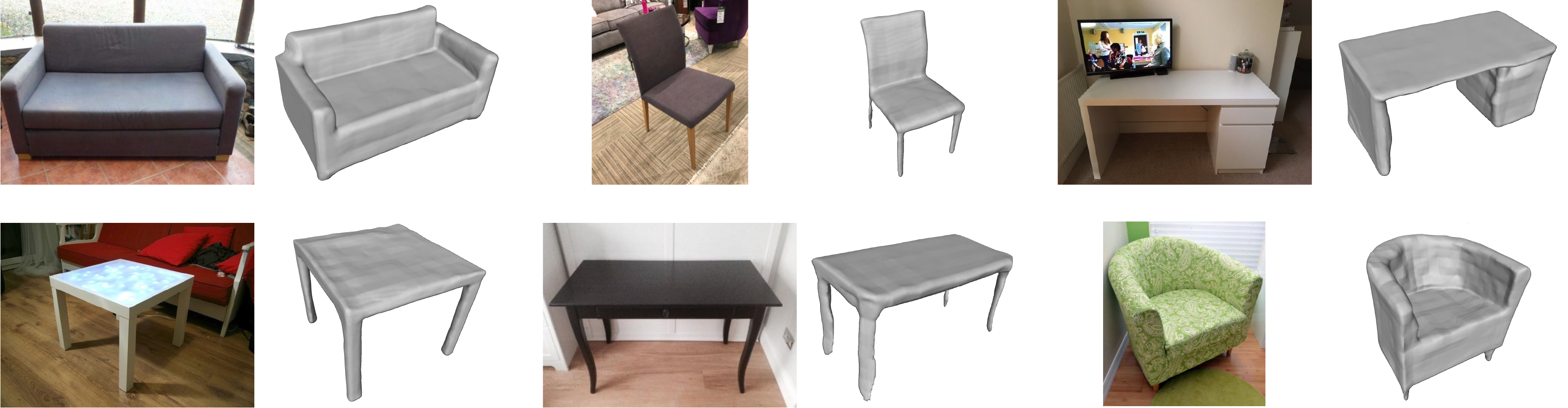}
\caption{
\label{fig:pix3dqual}
Single-view 3D reconstruction of our method on the real Pix3D dataset. The models were only trained on ShapeNet. \textit{Ground-truth masked} images were input to our model to obtain the reconstructions
}
\end{figure*}

\begin{figure*}
  \makebox[\textwidth][c]{\includegraphics[width=1.0\textwidth]{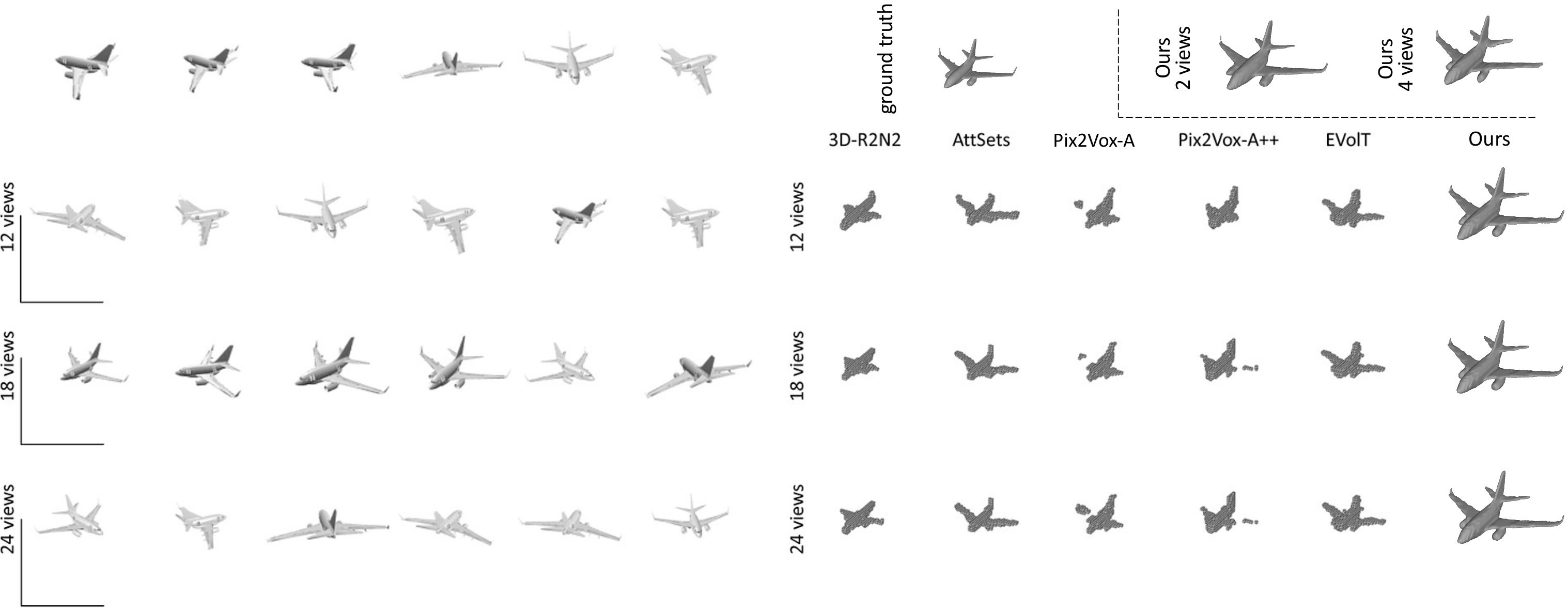}}%
  \caption{\label{fig:shapenetvox1} 3D reconstruction examples on the ShapeNet dataset compared to voxel based methods. We see that while the voxel based methods coarsely approximate the shape, \papertitle~reconstructs it with good detail even from a few views.}
\end{figure*}

\begin{figure*}
  \makebox[\textwidth][c]{\includegraphics[width=0.85\textwidth]{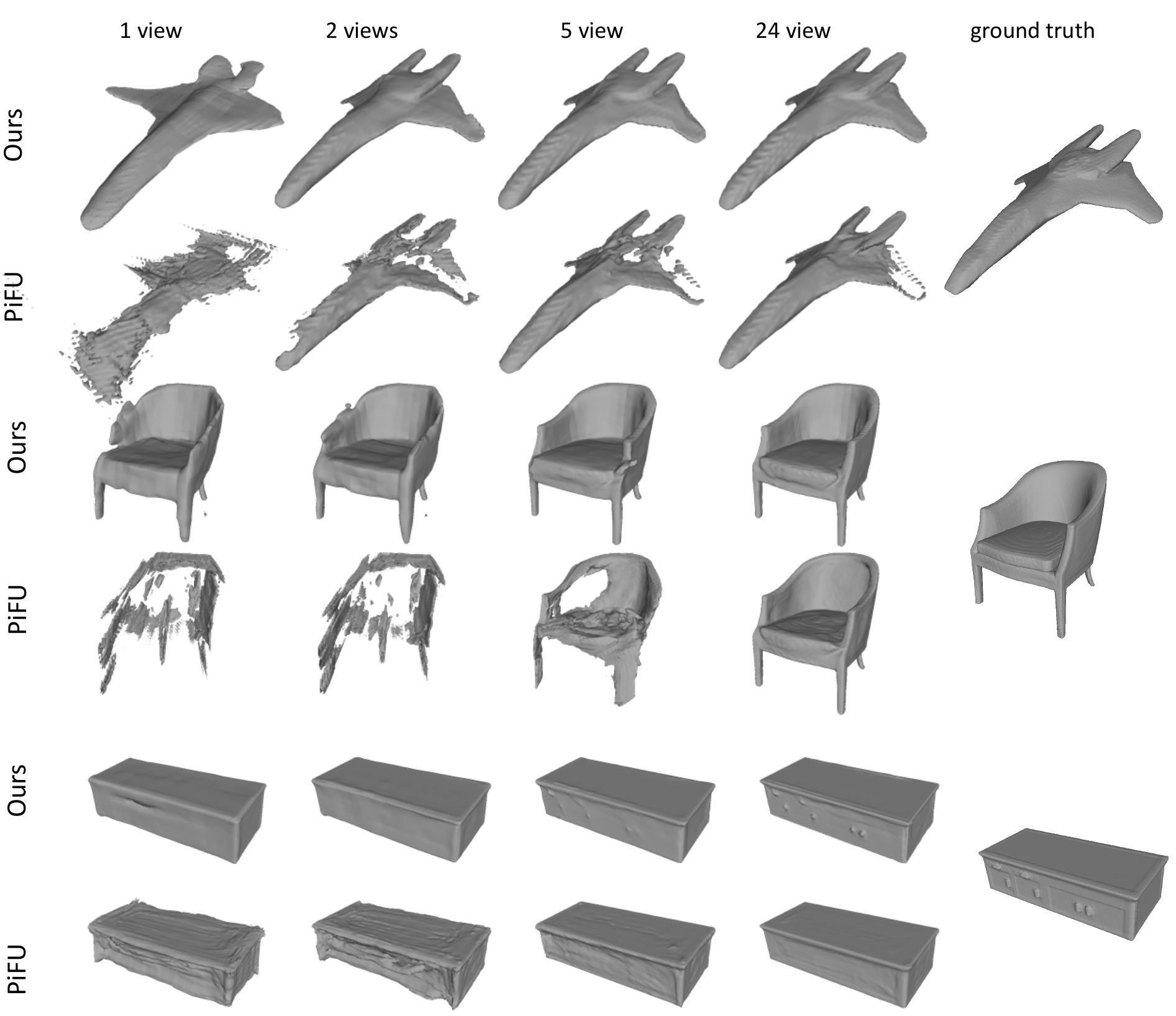}}%
  \caption{\label{fig:shapenetfig1} Some 3D reconstruction examples on the ShapeNet dataset. We see that \papertitle~attains better and more detailed reconstruction than PiFU from a lot less number of views.}
\end{figure*}

\begin{figure*}
  \makebox[\textwidth][c]{\includegraphics[width=1.0\textwidth]{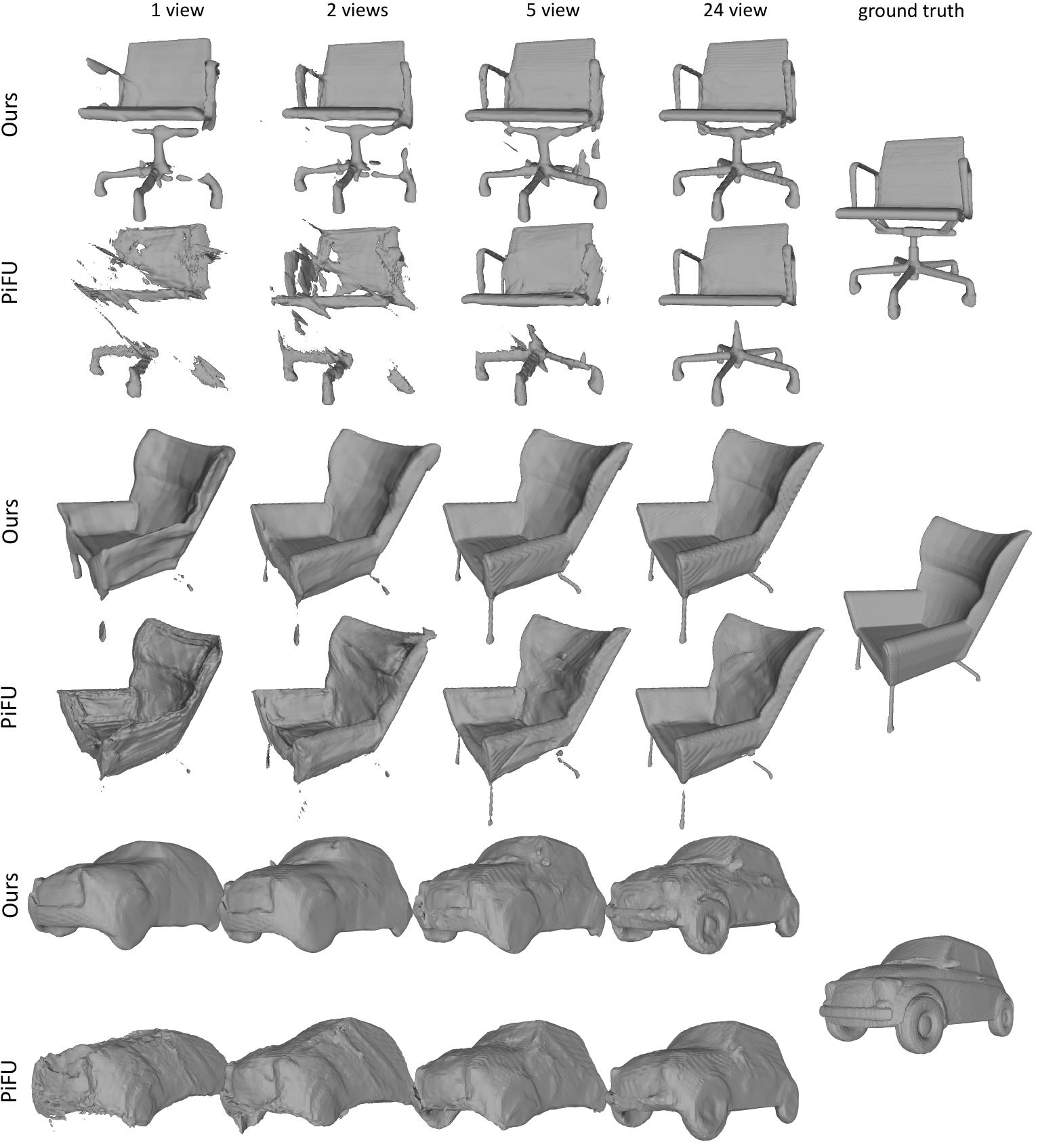}}%
  \caption{\label{fig:modelnetfig1} Some 3D reconstruction examples on the ModelNet dataset. We see that \papertitle~attains better reconstruction than PiFU from a lot less number of views.}
\end{figure*}

\begin{figure*}
  \makebox[\textwidth][c]{\includegraphics[width=1.0\textwidth]{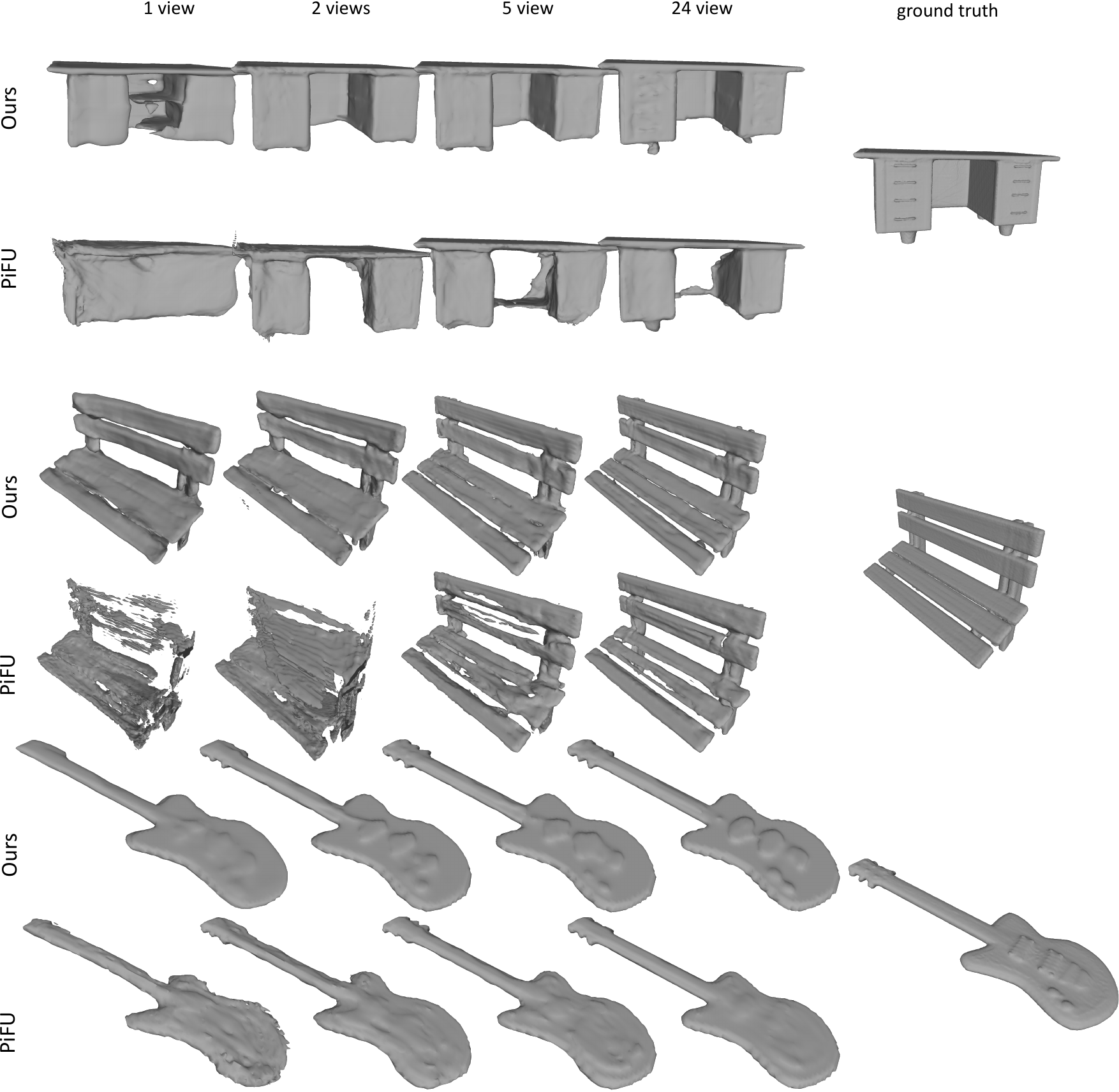}}%
  \caption{\label{fig:modelnetfig2} Some 3D reconstruction examples on the ModelNet dataset. We see that \papertitle~attains better reconstruction than PiFU from a lot less number of views.}
\end{figure*}

We qualitatively show synthetic-to-real adaptation of our method (\papertitle$_D$) by training it on ShapeNet and performing 3D reconstruction on Pix3D (Fig. \ref{fig:pix3dqual}).
This is also demonstrated quantitatively in Tab. \ref{singleviewreconstructionpix3d}.

We present more qualitative reconstruction results on the ShapeNet and ModelNet datatset.
We compare our full method \papertitle$_{Full}$ against voxel based methods in Fig.~\ref{fig:shapenetvox1}.
We see that the voxel based methods very coarsely approximate the shape, while even with 2 views, our method can capture the shape with good detail.
Fig.~\ref{fig:shapenetfig1} compares our method against PiFU.
We see that our method captures good detail even from a few views, while PiFU can suffer even with 24 views.
With the cabinet example in Fig.~\ref{fig:shapenetfig1}, we see that our method can capture finer detail with only 5 views, compared to PiFU using 24 views.
We also see similar comparative results on the ModelNet dataset in Fig.~\ref{fig:modelnetfig1}, ~\ref{fig:modelnetfig2}.

\clearpage

{\small
\bibliographystyle{ieee_fullname}
\bibliography{bibliography_jisan}
}

%% file: introduction.tex
\vspace{-4mm}
\section{Introduction}
\vspace{-2mm}
Recent technological advances in immersive technologies like virtual and augmented reality, robotics and autonomous systems, scene understanding, 3D CAD designing have rendered the task of dense 3D reconstruction an important problem to solve.
Recent ubiquity of depth sensors has inspired depth fusion for 3D reconstruction \eg \cite{izadi2011kinectfusion,niessner2013real,weder2020routedfusion,weder2020neuralfusion}.
The challenging problem of 3D reconstruction from RGB images has also been very popular \cite{snavely2006photo,Frahm:2010:BRC:1888089.1888117,mescheder2019occupancy,saito2019pifu}.
Traditional approaches like Structure From Motion (SfM) \cite{ozyecsil2017survey,schoenberger2016sfm} + Multi-View Stereo (MVS) \cite{lhuillier2005quasi,Frahm2010BuildingRO,goesele2007multi}, or Dense Simulteneous Localisation and Mapping (SLAM) \cite{kerl2013dense} often solved this problem using feature correspondences between images.
This becomes difficult with large self-occlusion and view-point variation.
In contrast, deep learning based techniques (\eg \cite{choy20163d,xie2019pix2vox,saito2019pifu,wang2021multi,runz2020frodo}) have achieved promising single or multi-view reconstruction performance utilizing appearance cues and contextual information.
Many of these works \cite{choy20163d,xie2019pix2vox,yang2020robust,wang2021multi} show improvements on voxel based 3D reconstruction.
However, the memory intensive nature of voxel representation forces these methods to output coarse reconstruction (often $32^3$ resolution).
Neural implicit representation approaches \cite{mescheder2019occupancy,saito2019pifu,xu2019disn,bautista2021generalization,runz2020frodo} have utilized neural network based continuous shape representation and have shown its ability to reconstruct objects with finer detail.

For multi-view reconstruction, 3D-R2N2 \cite{choy20163d} proposed recurrent neural network (RNN) based view aggregation.
However, RNNs cannot generate consistent results due to permutation variance \cite{vinyals2015order}.
They also suffer from long-term memory loss and high time complexity due to sequential nature.
Some works perform feature pooling along each 3D point (\eg mean pooling \cite{saito2019pifu} or max pooling \cite{xu2019disn}), or global feature mean pooling \cite{runz2020frodo} for multi-view fusion.
This can effectively ignore view association and may not capture different visibility conditions in different images.
Authors of \cite{xie2019pix2vox,yang2020robust} compute attention along each voxel separately, and fuse multi-view features using weighted attention.
Recently, EVolT \cite{wang2021multi} utilizes transformers \cite{vaswani2017attention} to establish pairwise view relationships and show that this allows better multi-view fusion for voxel based 3D reconstruction.
However, this method does not explicitly utilize 3D feature structure, and remains limited to a coarse $32^3$ resolution.

We propose \papertitle~- a framework that jointly utilizes a feature volume for semi-global and global context, and pixel-aligned features for fine local information towards neural implicit 3D reconstruction from single or multi-view imagery.
For each view, the feature volume provides 3D-structure-aware information, while the pixel-aligned features add fine-grained local information in a residual fashion.
For multi-view fusion, we jointly perform in an interleaved fashion - (1) 3D spatial reasoning along each feature volume, and (2) transformer-based pairwise multi-view reasoning along each local region of the feature volumes.
This allows our method to perform structurally coherent multi-view association at different scales.
We also perform pixel-aligned local feature fusion using a transformer-based network.
Using this representation and multi-view fusion, \papertitle~generates structurally coherent and high quality 3D reconstruction from any number of images (Fig. \ref{fig:mainfig}).
In summary, we make the following contributions:
\begin{itemize}[noitemsep,leftmargin=*]
    \item We propose \papertitle, a unified framework for single and multi-view neural implicit 3D reconstruction.
    \item \papertitle~utilizes both - 3D feature volume for global and semi-global context, and pixel-aligned features for fine local details - to generate high quality reconstruction.
    \item We design an interleaved 3D reasoning and multi-view reasoning network that uses pairwise view association for structurally coherent multi-view feature volume fusion.
    \item We propose to fuse multi-view pixel-aligned features, and augment them to the 3D feature volume representation.
    Using this rich representation, our method outperforms existing 3D reconstruction methods.
\end{itemize}

We demonstrate the efficacy of \papertitle~by comparing it to state-of-the-art (SOTA) reconstruction methods (Sec. \ref{experimentssection}).
\papertitle~outperforms or performs on-par other methods in all metrics for both single and multi-view reconstruction on the synthetic ShapeNet \cite{shapenet2015}, ModelNet \cite{wu20153d,SZB17a} and the real-world Pix3D \cite{sun2018pix3d} datasets.
\papertitle~attains superior reconstruction from only a few views, compared to other methods requiring a lot of views.

%% file: relatedworks.tex
\section{Related works}

\subsection{Neural Shape and Scene Representation}
\vspace{-2mm}
In the computer vision and graphics community, 3D shapes and scenes have historically been most commonly represented as mesh, pointclouds or voxels.
While compact, meshes are irregular in nature and are difficult to integrate into a deep learning frameworks \cite{michalkiewicz2019implicit}.
Pointclouds only provide sparse representation and are unable to model connectivity information.
Voxels are simple 3D analogy of 2D images, and represent scenes in fixed resolution grids.
However, they are very inefficient in representing high resolution scenes.
Neural implicit representations on the other hand usually represent 3D surfaces as a continuous decision boundary of a deep neural network classifier \cite{mescheder2019occupancy,chen2019learning}, or as the zero level set of a deep neural network regressor \cite{park2019deepsdf,michalkiewicz2019implicit}.
For these works, a 3D shape is usually represented as a latent code (\eg a latent vector).
A deep neural network based decoder conditioned on the latent code implicitly models the 3D surface - by estimating occupancy probability or Signed Distance Function (SDF) for query 3D points.
To scale neural implicit representations to larger scale, some works have proposed storing a grid of latent codes, each responsible for storing information about surfaces in a local neighborhood \cite{chabra2020deep,jiang2020local,genova2020local}.
In \papertitle, we propose to represent 3D objects by two components: (1) 3D feature volume which allows explicit 3D structure reasoning, instead of collapsing the whole representation onto one vector; and (2) pixel-aligned image features - which allows incorporating fine grained local information.

\subsection{Reconstruction from Single-view Image}
\vspace{-2mm}
Many single-view 3D reconstruction works have utilized 3D convolutional neural networks to generate 3D voxel based reconstruction \cite{choy20163d,tulsiani2017multi,xie2019pix2vox}.
While simple to represent, GPU memory requirement often limits these methods to relatively small $32^2$ grid resolution.
Meshes have also been considered as output representation \cite{kong2017using,groueix2018,wang2018pixel2mesh}.
These methods often suffer from problems like self-intersecting meshes, meshes with simple topology etc \cite{mescheder2019occupancy}.
Neural implicit representations have proven effective for this task by overcoming several such problems.
The authors of \cite{mescheder2019occupancy,chen2019learning} employ an encoder to obtain a global representation as a latent vector; which is decoded by a decoder into a 3D shape in its canonical orientation.
DmifNet \cite{dmifnetpaper} takes a similar approach and utilizes diverse representation from multi-branch information to improve reconstruction quality.
Despite the inherent advantage over voxel-based methods, a single global latent code limits these methods' ability to encode shape information with fine detail.
PiFU \cite{saito2019pifu} and DISN \cite{xu2019disn} show that instead of reducing image features into a global representation, utilizing image aligned features can perform the reconstruction with more detail.
Due to the lack of explicit 3D structure reasoning, these methods however cannot ensure 3D global shape consistency.
Geo-PiFU \cite{he2020geo} proposes to utilize a 3D U-Net \cite{resnet_2015} for additional 3D space reasoning.
However, \cite{saito2019pifu,he2020geo} were designed for detailed human shape reconstruction, and due to higher structural and geometric variation they do not perform well on man-made models (\eg~ShapeNet) \cite{li2021d2im}.
Compared to previous works, \papertitle~obtains high quality reconstruction by using both: a 3D feature volume to capture semi-global and global context; and pixel-aligned features to encode fine-grained information.
Unlike \cite{he2020geo}, \papertitle~is also naturally extendable to reconstruction from multi-view imagery.

\subsection{Reconstruction from Multi-view Images}
\vspace{-2mm}
Many approaches for multi-view shape reconstruction too have adopted voxel based reconstruction.
3D-R2N2 \cite{choy20163d} uses RNNs to perform multi-view reasoning.
But, RNNs suffer from being time consuming and not being permutation invariant.
Pix2Vox \cite{xie2019pix2vox} and AttSets \cite{yang2020robust} improve over this by computing per-voxel feature attention, and using weighted average to fuse them across views.
These methods compute attention independently for each view.
Hence, the attention computation for each view is completely unaware of the other observations.
Wang \etal \cite{wang2021multi} have shown that pairwise reasoning between multiple images using transformer \cite{vaswani2017attention} can lead to more effective information fusion towards improved voxel based reconstruction.
This method only utilized global image features, and not the fine pixel level details.
All these methods provide coarse voxel reconstruction of $32^3$ resolution.
Neural implicit representation based approaches also extend their method to 3D reconstruction from multi-view images.
These methods average multi-view features at each 3D point \cite{saito2019pifu,xu2019disn,bautista2021generalization}, without considering the 3D scene context in each observation.
3D43D \cite{bautista2021generalization} additionally uses pixel-aligned feature variance across views to approximate multi-view alignment cost.
FroDO \cite{runz2020frodo} simply mean pools global shape codes, which may putting equal emphasis on all observations.
In contrast, \papertitle~explicitly performs pairwise view association by interleaving 3D spatial reasoning and transformer based multi-view reasoning.
This jointly contextualizes 3D structure and multi-view observation.
\papertitle~also establishes pairwise view association for pixel-aligned features, leading to improved reconstruction.

%% file: methodology.tex
\vspace{-1mm}
\section{Method}
\vspace{-1mm}
We develop a unified 3D geometry reconstruction framework \papertitle, from posed single and multi-view RGB images.
\papertitle~proposes the construction and multi-view fusion of 3D feature volume for global and semi-global context, and pixel-aligned feature fusion for fine local detail.
It uses a combination of these features towards continuous neural implicit 3D reconstruction.

The outline of \papertitle~is illustrated is Fig. \ref{fig:mainfig}.
Given a number of posed RGB images $\{I^{(i)} \mid 1 \leq i \leq N\}$, we first use an image encoder to obtain feature map $g^{(i)}$ for each image $I^{(i)}$.
The encoder weights are shared between different images.
For each image, we then construct a 3D feature volume $G^{(i)} \in \mathbb{R}^{d \times d \times d \times c}$ in a uniform grid (Sec \ref{lfgridmaking}).
Features along each location of the feature volume corresponds to the scene constructs nearby.
We propose a joint spatial and multi-image reasoning network to establish spatial context, and fuse the feature volumes across views to obtain $F_g \in \mathbb{R}^{d \times d \times d \times c}$ (Sec \ref{interleavedunettransformer}).
$F_g$ gives us a 3D-structure-aware and multi-view aware fused feature representation.
Features $F_g(X)$ along any query 3D point $X$ can be computed using tri-linear interpolation.
To also utilize fine-grained local information, we project $X$ onto the image features, and fuse the pixel-aligned information across multiple views to obtain a representation $F_l(X)$ (Sec \ref{pixelalignedfeatures}).
Finally, the two features are combined in a residual fashion, followed by a decoder network to generate occupancy value $f(F_g(X) + F_l(X))$.
In this way, we continuously model the shape geometry, where the surface is a level set of $f$-

\vspace{-2mm}
\begin{equation}
    f(F_g(X) + F_l(X)) = \textbf{e} : \textbf{e} \in \mathbb{R}
\end{equation}

\papertitle~can perform reconstruction from single and multi-view images, since the transformer based cross-view fusion components can fuse features of a set of any size.
For mesh extraction, we evaluate $f$ in a dense grid, followed by the Marching Cube algorithm \cite{lorensen1987marching,lewiner2003efficient}.
We train \papertitle~by sampling random 3D points and penalizing their predicted occupancy w.r.t. the ground-truth occupancy.
\vspace{-1mm}
\subsection{3D Feature Volume Construction and Fusion}
\vspace{-1mm}
\papertitle~creates a uniform 3D feature volume for each image and performs spatial reasoning to establish 3D-structure-aware context.
Each location in the 3D feature volume effectively captures nearby shape structure.
The module also performs pairwise view association to share information in-between views.
The feature volumes are finally fused together to obtain 3D-structure-aware and multi-view-aware feature representation.

\begin{figure*}[t!]
\centering
\vspace{-3mm}
\includegraphics[width=.95\linewidth]{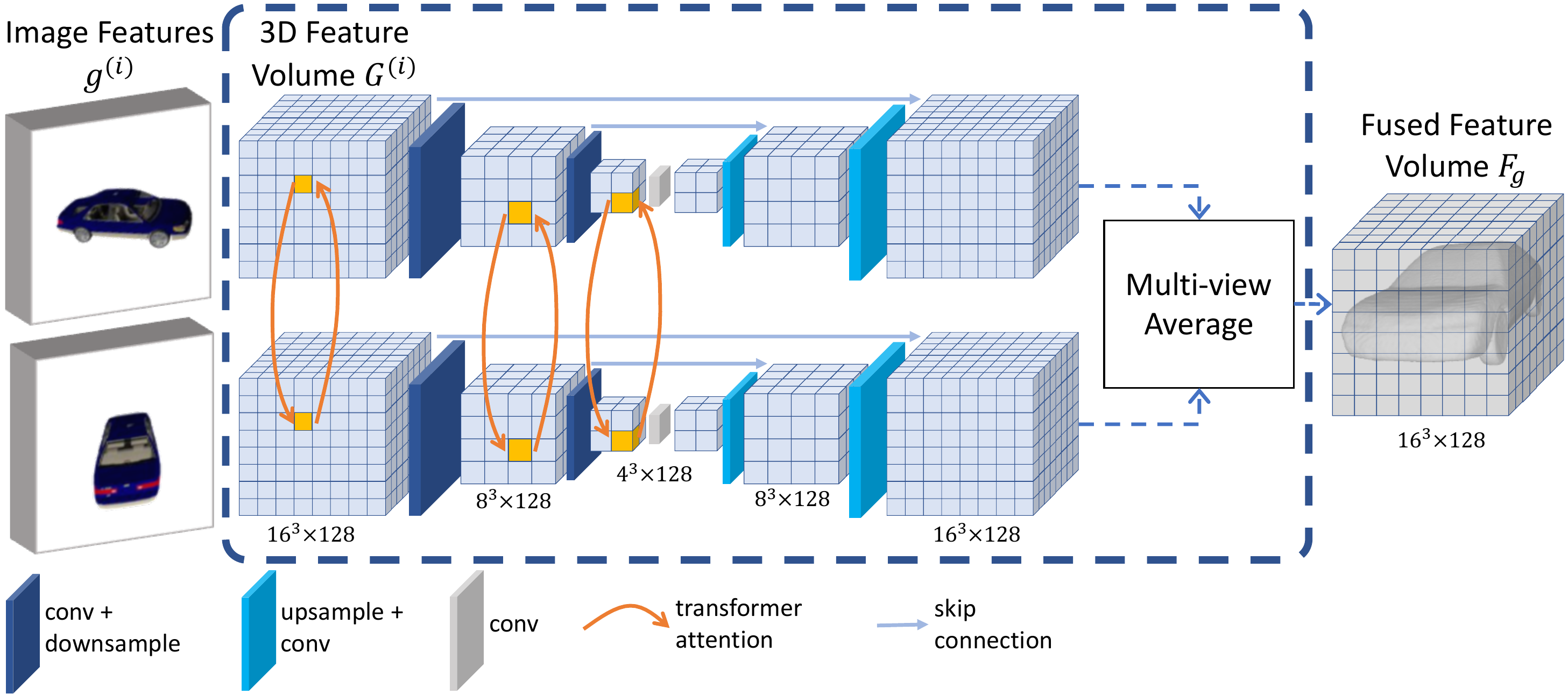}
\vspace{-2mm}
\caption{\label{fig:fusionfig} Interleaved 3D-UNet and multi-view feature association using transformers. A transformer attention is applied after each U-Net encoder block, for the same 3D feature volume location across different views. For illustration simplicity, we show the pairwise view association between only two views here. For $N$ such views, there will be $\binom{N}{2}$ such pairs for view association. Performing multi-view reasoning at different scales in this fashion allows our network to extract 3D-structure-aware and multi-view aware features. \vspace{-5mm}}
\end{figure*}

\vspace{-1mm}
\subsubsection{Creating 3D Feature Volume}
\vspace{-1mm}
\label{lfgridmaking}
We create a 3D feature volume for each image $I^{(i)}$.
First, we uniformly divide the 3D volume in a $d\times d \times d$ grid.
We sample feature for each grid cell by projecting the cell center onto feature map $g^{(i)}$ and sampling bi-linearly.
As multiple 3D points can project to the same feature map location, we also extract the depth of these cell centers.
We then construct the 3D feature volume $G^{(i)} \in \mathbb{R}^{d \times d \times d \times c}$; where each cell encodes a concatenation of the sampled features and the positional encoded \cite{mildenhall2020nerf} depth, a total of $c$ channels.

\vspace{-1mm}
\subsubsection{3D Reasoning and Feature Volume Fusion}
\vspace{-1mm}
\label{interleavedunettransformer}
This sub-module takes the 3D feature volumes $\left\{ G^{(i)} \right\}$ for all images, performs both spatial reasoning and cross-image reasoning, and fuses the multi-view representation into one 3D-structure-aware and multi-view-aware feature volume representation $F_g \in \mathbb{R}^{d \times d \times d \times c}$.
For all multi-view reasoning, we utilize attention along each 3D feature volume location across different views.
This allows different observations to share information about the same part of the 3D scene regarding visibility and occlusion.
For 3D spatial reasoning and multi-view reasoning, we propose several architectures.
All of them model multi-view reasoning as a permutation invariant set operation. \\
\textbf{A. 3D U-Net + Multi-image AttSets Attention:}
a 3D UNet \cite{DBLP:journals/corr/RonnebergerFB15} is applied on each feature volume $G^{(i)}$ for spatial reasoning.
A sequence 3 of AttSet residual blocks \cite{yang2020robust} then compute attention along 3D feature volume locations in different views.
Finally, attention weighted feature summation across views creates the fused feature volume $F_g$. \\
\textbf{B. 3D U-Net + Multi-image Transformer Attention:}
similar to before - a 3D UNet is first applied on each feature volume $G^{(i)}$ to perform spatial reasoning.
Then a sequence of 3 transformer self-attention layers \cite{vaswani2017attention} are applied along each 3D feature volume location across different views, and finally averaged across views to obtain the fused feature volume $F_g$.
To keep the computed attention permutation invariant to view order, we do not use any positional encoder for the transformer.
This architecture explicitly performs pairwise view association using transformers, while the former computes attention for each view independently.\\
\textbf{C. Axial Attention:} we construct a feature tensor combining all 3D feature volumes $\textbf{G} \in \mathbb{R}^{N \times d \times d \times d \times c}$, and apply transformer attention steps in dimension 1, 2, 3 and 4 alternately.
This architecture is inspired by \cite{ho2019axial,bertasius2021space}.
While a full attention over multi-image and spatial axes would impractically require evaluating $\mathcal{O}\left(\left(Nd^3\right)^2\right)=\mathcal{O}\left(N^2d^6\right)$ attention relations, an axial attention block evaluates only $\mathcal{O}\left(N^2+3d^2\right)$ relations.
For the multi-image dimension (axis 1) we ignore positional encoding for permutation invariance.
We apply regular sequence attention for the spatial dimensions.
We apply 3 such blocks, and then average them across views to obtain the fused feature volume $F_g$.

\noindent \textbf{D. Interleaved 3D U-Net and Multi-image Attention:}
to allow joint multi-image and spatial reasoning at different scales, we propose a network that interleaves - 3D U-Net on each 3D feature volume and multi-image attention across different feature volumes (Fig. \ref{fig:fusionfig}).
After each U-Net encoder block, we apply a multi-view attention block (consisting of 2 transformer self-attention layers) along each 3D feature volume location across different views - to share information between different views at that scale.
The attention performs pairwise view association to implicitly reason about the visibility and occlusion conditions of different scene parts in each view.
It reasons locally at the earlier stages, and reasons semi-globally at the the later stages.
Altogether, the multi-view attention blocks operate at different scales of the feature volumes.
Hence, this architecture has the advantage over the other proposed architectures in being able to utilize local and semi-global context across all image views.
After applying the interleaved 3D U-Net and multi-image attention, each feature volume is 3D-structure-aware and all view aware.
Finally, we average the features along different views to obtain fused feature volume $F_g$.

\vspace{-1mm}
\subsection{Pixel-aligned Local Features Fusion}
\vspace{-1mm}
\label{pixelalignedfeatures}
We extract pixel-aligned features from each feature map for any 3D point $X$ to utilize local fine-detailed information.
Similar to \cite{saito2019pifu}, we project 3D point $X$ onto each 2D feature map $g^{(i)}$ and bi-linearly sample the image features.
We also extract the depth of $X$ in each image $I^{(i)}$.
Unlike \cite{saito2019pifu} that just mean pools features, we apply a sequence of 3 transformer self-attention layers to perform cross-image reasoning for $X$.
This allows multi-view awareness of the pixel-aligned features from each view - before they are fused together.
Finally, we perform feature averaging to obtain fused fine-grained pixel-aligned features $F_l(X)$.
\vspace{-1mm}
\subsection{Shape Reconstruction}
\vspace{-1mm}
\label{finalreconstruction}
For any 3D point $X$, we tri-linearly sample the fused feature volume $F_g$ to obtain $F_g(X)$.
This is a 3D-structure-aware and multi-view aware feature representation for $X$.
We residually add the pixel-aligned features $F_l(X)$ to this representation.
Finally, this feature is passed through a small MLP-based decoder to predict the occupancy value along point $X$.
The reconstruction is continuous, since the network can be evaluated at any 3D point.
To obtain a mesh reconstruction - we predict occupancy on a 3D grid of points, followed by Marching Cubes \cite{lewiner2003efficient}.
Alternatively, MISE \cite{mescheder2019occupancy} could be used for faster mesh extraction.

%% file: experiments.tex
\vspace{-1mm}
\section{Experiments}
\vspace{-1mm}
\label{experimentssection}
\subsection{Implementation Detail}
\vspace{-1mm}
We use a stacked hourglass architecture \cite{newell2016stacked} similar to \cite{saito2019pifu}, to extract image features.
For the feature volumes, we set the spatial dimension $d = 16$.
All transformer layers use 8 heads and feed-forward layer dimension of 128.
We use the DeepSDF \cite{park2019deepsdf} decoder for occupancy decoding.
We remove the 3D query point conditioning, since $F_g(X) + F_l(X)$ already corresponds to features specific to $X$.
More detail on our network can be found in the supplementary.

At each training iteration, we randomly sample $K$ images ($1 \leq K \leq 8$) per object reconstruction.
The batch size is set at $\left \lfloor \frac{16}{K} \right \rfloor$ objects.
For each object, $2048$ points are randomly sampled using a combination of random uniform and close-to-surface samples (1:5 ratio).
The predicted occupancies along these points are penalized using cross-entropy loss.
We have found it beneficial to train the full network in two stages: (1) we train only using the fused 3D feature volume $F_g$ for the reconstruction. Then, (2) we also add the pixel-aligned fused local features and train the whole system.
This led to overall better reconstruction quality compared to training using both features from the beginning.

To extract mesh, we evaluate the occupancy in a uniform $128^3$ grid, followed by Marching Cube at level-set $\textbf{e}$ = 0.43.
More training details are included in the supplementary.

\input{tables/singleviewshapenetlarge}

\input{tables/multiviewshapenet}

\vspace{-1mm}
\subsection{Quantitative Results}
\vspace{-1mm}

\noindent \textbf{Metrics:}
To measure the reconstruction performance, we evaluate volumetric IoU, Chamfer-$L1$ and normal-consistency metrics - similar to \cite{mescheder2019occupancy}.
We evaluate Intersection-over-Union (IoU) of the reconstructed shape and the ground-truth mesh, which is the ratio of intersection of the volume to their union.
An unbiased estimate of the volumes are computed by randomly sampling 100K points from the bounding volume and determining if points lie within or outside the predicted/ground-truth mesh.
The Chamfer-$L1$ metric is computed as the mean of \textit{accuracy} and \textit{completeness}.
Accuracy is the mean distance of points on the reconstructed surface to their nearest neighbor in the ground-truth surface.
This is estimated using 100K randomly sampled points from both surfaces.
Completeness is calculated in the same way, but in the opposite direction.
Same as \cite{fan2017point}, we use $1/10$ times the maximal edge length of each object's bounding box as 1 unit.
To also measure how well the reconstruction captures higher order information, normal consistency is calculated as the mean absolute dot product of the normals in one mesh and the normals at the nearest neighbors in the other mesh.
We also show that the proposed method has favorable 3D reconstruction runtime against other methods (in the supplementary).

\vspace{1mm}
\noindent \textbf{Datasets:}
We evaluate the reconstruction performance on ShapeNet \cite{shapenet2015} subset of 3D-R2N2 \cite{choy20163d}.
It consists of 43,783 3D shapes of 13 different categories.
We use the same train/test split and image renderings as 3D-R2N2.
We also evaluate on the ModelNet \cite{wu20153d} aligned subset by Sedaghat \etal \cite{SZB17a}.
This contains 12,311 3D models of 40 categories.
Similar to 3D-R2N2, we create 24 renders around each object - with random camera elevation in $[15^{\circ}, 60^{\circ}]$.
We use codes provided by \cite{Stutz_2018_CVPR,choy20163d} for generating watertight ground-truth meshes and mesh occupancy computation.
We also evaluate the synthetic training to real-world single view reconstruction adaptation - by training on ShapeNet, and evaluating them on Pix3D \cite{sun2018pix3d}.

\vspace{1mm}
\noindent \textbf{Single-view Reconstruction:}
For single-view 3D reconstruction, we compare the proposed \papertitle~to 3D-R2N2 \cite{choy20163d}, Pixel2Mesh \cite{wang2018pixel2mesh}, AtlasNet \cite{groueix2018}, ONet \cite{mescheder2019occupancy}, DmifNet \cite{dmifnetpaper}, CoReNet \cite{corenetpaper}, 3D43D \cite{bautista2021generalization} and Ray-ONet \cite{bian2021rayonet} (Tab.~\ref{singleviewreconstructiontablelarge}).
The performance metrics for the first three methods are taken from \cite{mescheder2019occupancy}, while the CoReNet numbers are taken from \cite{bian2021rayonet}.
The rest are taken from the respective papers.
The training protocol of ONet \cite{mescheder2019occupancy} was used for all baseline methods.
All methods in Tab.~\ref{singleviewreconstructiontablelarge} are evaluated against the ground-truth 3D mesh.
Here, `Ours' refers to our complete model \papertitle$_{Full}$.
We use both - fused 3D feature volume (architecture \textit{D. Interleaved 3D U-Net and multi-image attention}), and fused pixel-aligned local features for this reconstruction.
Our method achieves better volumetric IoU (higher is better) compared to other methods across most object categories by utilizing both 3D-structure-aware and pixel-aligned feature representations.
Our method (mean IoU/mIoU 0.664) also greatly outperforms DISN (mIoU 0.594) and PiFU (mIoU 0.483).
Tab.~\ref{singleviewreconstructiontablelarge} also shows that our method achieves better or competitive Chamfer-$L1$ (lower is better) score against other methods in most categories, while outperforming all methods in the mean Chamefer-$L1$ metric.
\papertitle~also attains higher normal consistency against baseline methods in most categories.
While voxel based methods like Pix2Vox \cite{xie2019pix2vox}, EVolT \cite{wang2021multi}, Mem3D \cite{mem3dpaper} have a mIoU upper-bound of $0.631$ due to their coarse $32^3$ resolution, VPFusion achieves mIoU of $0.664$ only from one view.
We also experiment on adaptation to real data by training on ShapeNet and testing on single-view reconstruction dataset Pix3D.
The results are in the supplementary.
\input{tables/multiviewmodelnet}

\noindent \textbf{Multi-view Reconstruction:}
Tab.~\ref{tab:multiviewtable} presents the multi-view reconstruction performance on ShapeNet.
Many multi-view reconstruction techniques (\eg 3D-R2N2 \cite{choy20163d}, AttSets \cite{yang2020robust}, Pix2Vox \cite{xie2019pix2vox}, EVolT \cite{wang2021multi}) generate voxel reconstruction.
Part (a) of Tab.~\ref{tab:multiviewtable} shows the mean IoU of voxel based methods \textit{measured against ground-truth voxels} (taken from \cite{wang2021multi}).
Part (b) shows the mean IoU of different methods \textit{measured against ground-truth meshes}.
All of the voxel based methods in part (a) are limited to $32^3$ resolution.
As a result, their performance upper bound (using ground-truth voxels) is $0.631$ when compared against ground-truth meshes (part (b) - first row).
We also present the performance of implicit representation works ONet$^\dagger$ \cite{mescheder2019occupancy}, 3D43D \cite{bautista2021generalization} and PiFU \cite{saito2019pifu}.
ONet$^\dagger$ in this case was extended to multi-view case \cite{bautista2021generalization} by mean pooling the conditional features across views at inference.

We evaluate 5 different variants of \papertitle~for this task.
\papertitle$_{A}$ through \papertitle$_{D}$ refer to our models using only multi-view fused 3D feature volume (described in Sec. \ref{interleavedunettransformer}), without using the pixel-aligned local features.
We see that \papertitle$_{B}$ achieves better mean IoU than \papertitle$_{A}$ for multi-view reconstruction.
This indicates the advantage of establishing pairwise image association using transformers over the approach by AttSets \cite{yang2020robust}.
\papertitle$_{C}$ shows that instead of applying 3D reasoning and multi-view attention one after another, it is beneficial to perform these operations alternately.
However, \papertitle$_{D}$ further outperforms that using proposed interleaved 3D-UNet and multi-view attention architecture.
Our complete model is \papertitle$_{Full}$, which uses the 3D feature volume as in \papertitle$_{D}$, and also utilizes multi-view fused pixel-aligned features $F_l(.)$.
This achieves the best performance among all variants, demonstrating the efficacy of the proposed model over all SOTA models and different variants of \papertitle.
Especially, \papertitle$_{Full}$ attains a mean IoU of 0.816 using only 7 views, which is higher than that of other methods using upto 24 views.

Tab.~\ref{tab:multiviewmodelnettable} shows the multi-view reconstruction performance of the most competitive baseline PiFU and the most competitive variants of \papertitle~on ModelNet.
We again observe similar result, where our method achieves similar IoU using 12 views compared to PiFU using upto 24 views.

\vspace{-1mm}
\subsection{Qualitative Results}
\begin{figure*}[h!]
  \vspace{-5mm}
  \makebox[\textwidth][c]{\includegraphics[width=1.05\textwidth]{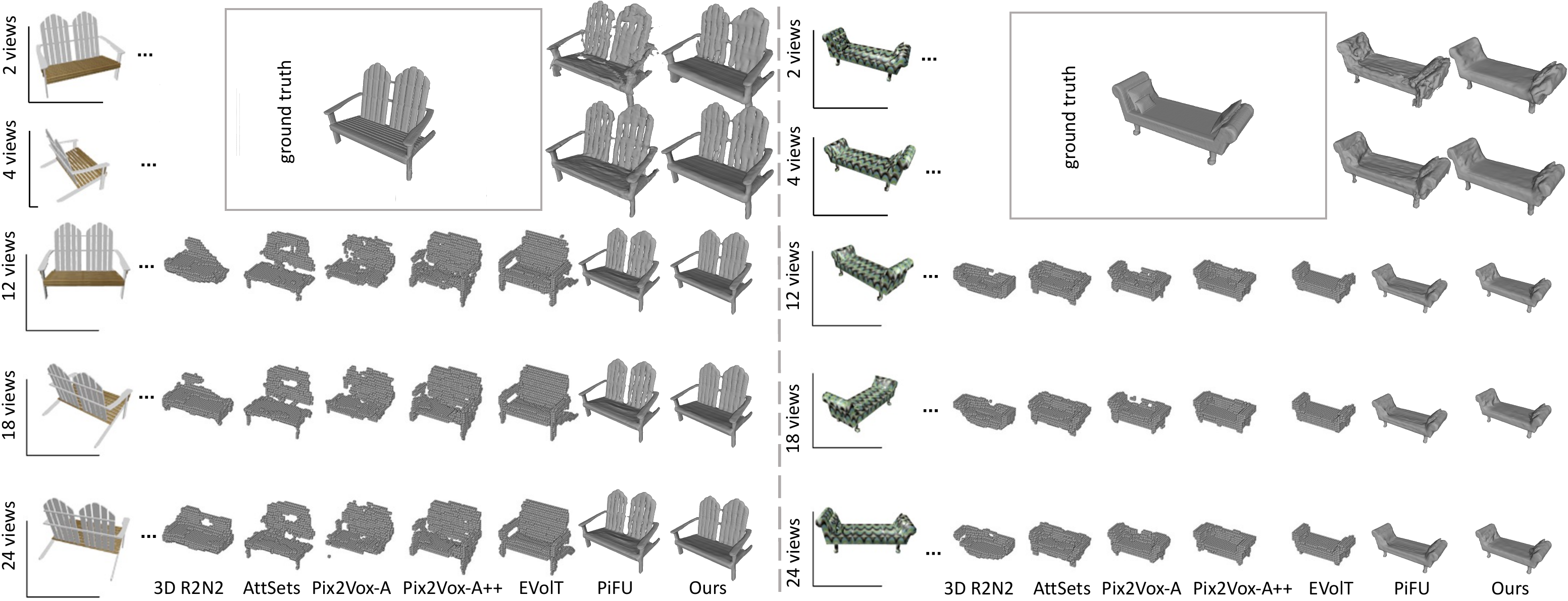}}%
  \vspace{-2mm}
  \caption{\label{fig:qualresult} Some 3D reconstruction examples on the ShapeNet dataset by different methods. While other methods require 12, 18 or 24 views to obtain a reasonable reconstruction, \papertitle~can already achieve higher reconstruction quality with 4 views. Given additional views, our method can further refine the reconstruction quality. While PiFU performs well on large number of views, our approach can reconstruct well even from sparse views. Please zoom in to see better reconstruction detail. \vspace{-2mm}}
\end{figure*}

\begin{figure*}[!h]
\centering
\includegraphics[width=0.83\linewidth]{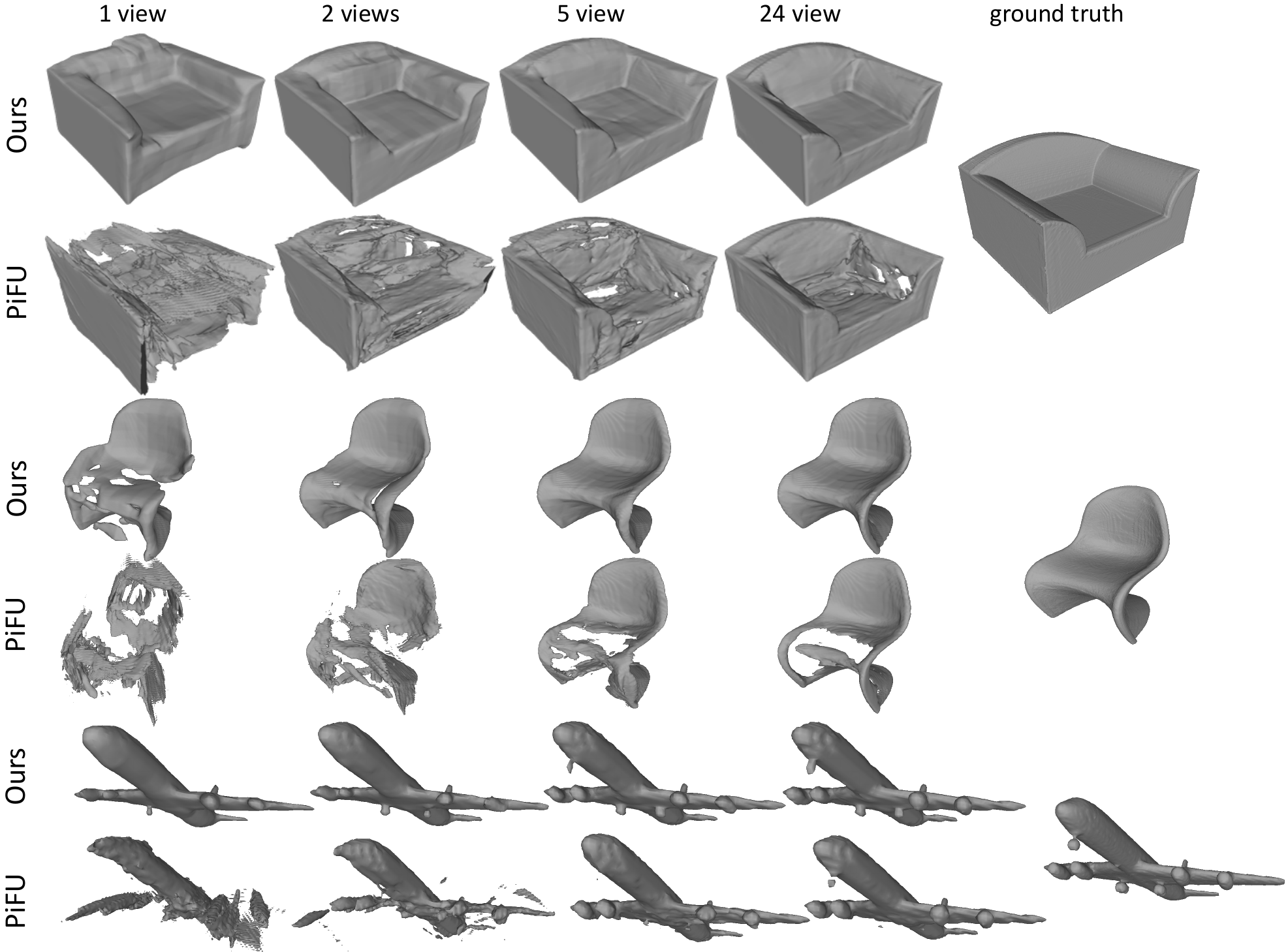}
\caption{
\label{fig:vspifu} Compared to PiFU, \papertitle~attains higher-quality reconstruction from much fewer number of images by utilizing the proposed joint 3D feature volume fusion and pixel-aligned local feature fusion.
Please zoom in to see better reconstruction detail. \vspace{-5mm}
}
\end{figure*}

\vspace{-1.5mm}
We qualitatively compare \papertitle~to 3D-R2N2, AttSets, Pix2Vox-A, Pix2Vox-A++, EVolT and PiFU.
Fig.~\ref{fig:qualresult} presents some multi-view reconstruction performance on ShapeNet for 12, 18 and 24 views respectively.
The figures for voxel based methods are taken from \cite{wang2021multi}.
Voxel based methods suffer from low resolution output, possibly even showing signs of class based over-fitting as indicated in the bench example.
PiFU performs well given a lot of view redundancy, but it suffers on few-view reconstruction.
We see that compared to all other methods, \papertitle$_{Full}$ can reconstruct the object geometry very faithfully from a few views.
\papertitle$_{Full}$ obtains good reconstruction performance even with 2 or 4 views - indicating the efficacy of our fusion architecture.
Given more number of views, our method successively improves the reconstruction quality further.
Fig.~\ref{fig:vspifu} shows in more detail how our method achieves higher reconstruction quality on ShapeNet compared to PiFU using much fewer number of images.
Some synthetic-to-real adaptation examples (by \papertitle$_D$) are in Fig. \ref{fig:pix3dqual}.
More qualitative results on ShapeNet, ModelNet and Pix3D datasets are presented in the supplementary.

\begin{figure}[!t]
\centering
\vspace{-1.5mm}
\includegraphics[width=0.96\linewidth]{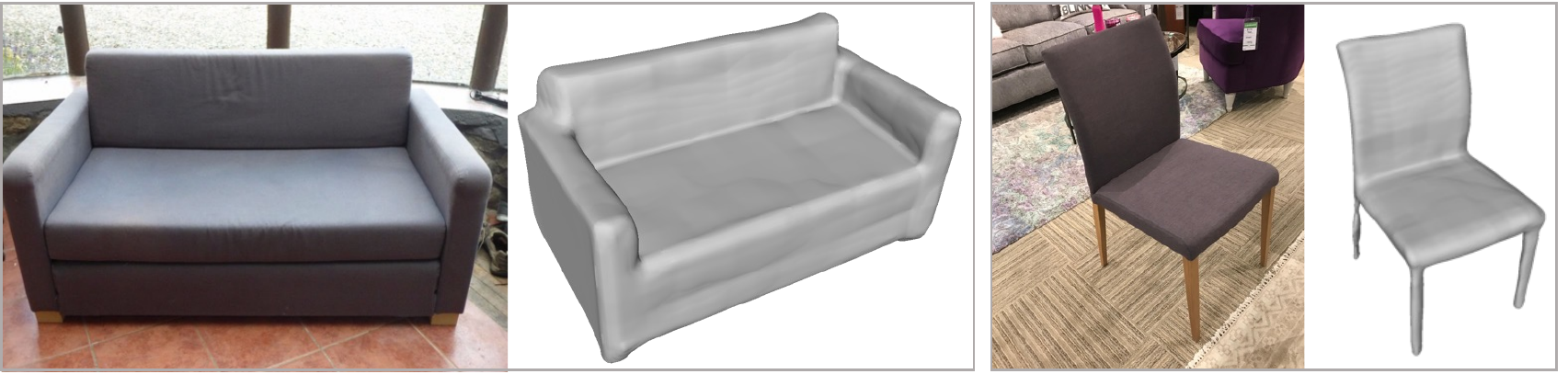}
\vspace{-2mm}
\caption{
\label{fig:pix3dqual}
Synthetic ShapeNet training to real-world adaptation examples on the ground-truth masked single-view Pix3D images.
\vspace{-5mm}
}
\end{figure}

\vspace{-1mm}
\subsection{Ablation}
\vspace{-1mm}
For ablative study, we study the efficacy of our choice of feature representation and architecture design.
Compared to PiFU \cite{saito2019pifu} which only utilizes pixel-aligned local features, we find the construction of 3D feature volume and multi-view reasoning along each feature volume location highly effective, as indicated in Tab.~\ref{tab:multiviewtable}, \ref{tab:multiviewmodelnettable}.
For multi-view fusion, utilizing transformers to establish pair-wise association between images (\papertitle$_{B}$) leads to a more effective multi-view reasoning compared to the AttSets approach of computing them independently (\papertitle$_{A}$).
We find the interleaved 3D U-Net and multi-view transformer architecture (\papertitle$_{D}$) to be very beneficial towards 3D-structure-aware and multi-view aware feature extraction.
This is attested by the great reconstruction performance improvement over \papertitle$_{B}$, which applies the multi-view transformer only after the 3D-UNet.
Finally, we also observe the importance of pixel-aligned local features.
Augmenting this feature to the 3D feature volume residually to build \papertitle$_{Full}$ further improves the reconstruction quality.

%% file: tables/singleviewshapenetlarge.tex
\begin{table*}[!ht]
\centering
\vspace{-3mm}
\resizebox{1.0\linewidth}{!}{
\begin{tabular}
{c|c|cccccccccccccc}
\toprule
 Metric & Method & plane & bench & cabinet & car & chair & display & lamp & speaker & rifle & sofa & table & phone & vessel & mean \\
\toprule
\multirow{7}{*}{\centering IoU $\uparrow$} & 3D-R2N2 & 0.426 & 0.373 & 0.667 & 0.661 & 0.439 & 0.440 & 0.281 & 0.611 & 0.375 & 0.626 & 0.420 & 0.611 & 0.482 & 0.493 \\
 & Pix2Mesh & 0.420 & 0.323 & 0.664 & 0.552 & 0.396 & 0.490 & 0.323 & 0.599 & 0.402 & 0.613 & 0.395 & 0.661 & 0.397 & 0.480 \\
 & AtlasNet & - & - & - & - & - & - & - & - & - & - & - & - & - & - \\
 & ONet & 0.571 & 0.485 & 0.733 & 0.737 & 0.501 & 0.471 & 0.371 & 0.647 & 0.474 & 0.680 & 0.506 & 0.720 & 0.530 & 0.571 \\
 & DmifNet & 0.603 & 0.512 & 0.753 & 0.758 & 0.542 & 0.560 & 0.416 & 0.675 & 0.493 & 0.701 & 0.550 & 0.750 & 0.574 & 0.607 \\
 & CoReNet & 0.543 & 0.431 & 0.715 & 0.718 & 0.512 & 0.557 & 0.430 & 0.686 & 0.589 & 0.686 & 0.527 & 0.714 & 0.562 & 0.590 \\
 & 3D43D & 0.571 & 0.502 & 0.761 & 0.741 & 0.564 & 0.548 & 0.453 & \textbf{0.729} & 0.529 & 0.718 & 0.574 & 0.740 & 0.588 & 0.621 \\
 & Ray-ONet & 0.574 & 0.521 & 0.763 & 0.755 & 0.567 & 0.601 & 0.467 & 0.726 & 0.594 & 0.726 & 0.578 & 0.757 & 0.598 & 0.633 \\
 & Ours & \textbf{0.680} & \textbf{0.551} & \textbf{0.782} & \textbf{0.767} & \textbf{0.582} & \textbf{0.633} & \textbf{0.510} & 0.720 & \textbf{0.670} & \textbf{0.742} & \textbf{0.590} & \textbf{0.775} & \textbf{0.626} & \textbf{0.664} \\
 \midrule
\multirow{7}{*}{\centering Chamfer-$L1 \downarrow$} & 3D-R2N2 & 0.227 & 0.194 & 0.217 & 0.213 & 0.270 & 0.314 & 0.778 & 0.318 & 0.183 & 0.229 & 0.239 & 0.195 & 0.238 & 0.278 \\
 & Pix2Mesh & 0.187 & 0.201 & 0.196 & 0.180 & 0.265 & 0.239 & 0.308 & 0.285 & 0.164 & 0.212 & 0.218 & 0.149 & 0.212 & 0.216 \\
 & AtlasNet & 0.104 & 0.138 & 0.175 & 0.141 & 0.209 & 0.198 & 0.305 & 0.245 & 0.115 & 0.177 & 0.190 & 0.128 & 0.151 & 0.175 \\
 & ONet & 0.147 & 0.155 & 0.167 & 0.159 & 0.228 & 0.278 & 0.479 & 0.300 & 0.141 & 0.194 & 0.189 & 0.140 & 0.218 & 0.215 \\
 & DmifNet & 0.131 & 0.141 & 0.149 & 0.142 & 0.203 & 0.220 & 0.351 & 0.263 & 0.135 & 0.181 & 0.173 & 0.124 & 0.189 & 0.185 \\
 & CoReNet & 0.135 & 0.157 & 0.181 & 0.173 & 0.216 & 0.213 & 0.271 & 0.263 & 0.091 & 0.176 & 0.170 & 0.128 & 0.216 & 0.184 \\
  & 3D43D & 0.096 & \textbf{0.112} & \textbf{0.119} & \textbf{0.122} & 0.193 & \textbf{0.166} & 0.561 & 0.229 & 0.248 & \textbf{0.125} & 0.146 & \textbf{0.107} & 0.175 & 0.184 \\
 & Ray-ONet & 0.121 & 0.124 & 0.140 & 0.145 & \textbf{0.178} & 0.183 & 0.238 & \textbf{0.204} & 0.094 & 0.151 & \textbf{0.140} & 0.109 & 0.163 & 0.153 \\
 & Ours & \textbf{0.093} & 0.124 & 0.128 & 0.128 & 0.183 & 0.176 & \textbf{0.236} & 0.208 & \textbf{0.077} & 0.141 & 0.150 & 0.113 & \textbf{0.148} & \textbf{0.147} \\
 \midrule
\multirow{7}{*}{\centering \begin{tabular}{c} Normal \\ Consistency $\uparrow$ \end{tabular}} & 3D-R2N2 & 0.629 & 0.678 & 0.782 & 0.714 & 0.663 & 0.720 & 0.560 & 0.711 & 0.670 & 0.731 & 0.732 & 0.817 & 0.629 & 0.695 \\
 & Pix2Mesh & 0.759 & 0.732 & 0.834 & 0.756 & 0.746 & 0.830 & 0.666 & 0.782 & 0.718 & 0.820 & 0.784 & 0.907 & 0.699 & 0.772 \\
 & AtlasNet & 0.836 & 0.779 & 0.850 & 0.836 & 0.791 & 0.858 & 0.694 & 0.825 & 0.725 & 0.840 & 0.832 & 0.923 & 0.756 & 0.811 \\
 & ONet & 0.840 & 0.813 & 0.879 & 0.852 & 0.823 & 0.854 & 0.731 & 0.832 & 0.766 & 0.863 & 0.858 & 0.935 & 0.794 & 0.834 \\
 & DmifNet & 0.853 & 0.821 & 0.885 & \textbf{0.857} & 0.835 & 0.872 & 0.758 & 0.847 & 0.781 & 0.873 & 0.868 & 0.936 & 0.808 & 0.846 \\
 & CoReNet & 0.752 & 0.742 & 0.814 & 0.771 & 0.764 & 0.815 & 0.713 & 0.786 & 0.762 & 0.801 & 0.790 & 0.870 & 0.722 & 0.777 \\
 & 3D43D & 0.825 & 0.809 & 0.886 & 0.844 & 0.832 & 0.883 & 0.766 & \textbf{0.868} & 0.798 & 0.875 & 0.864 & 0.935 & 0.799 & 0.845 \\
 & Ray-ONet & 0.830 & 0.828 & 0.881 & 0.849 & \textbf{0.844} & 0.883 & 0.780 & 0.866 & 0.837 & 0.881 & \textbf{0.873} & 0.929 & 0.815 & 0.854 \\
 & Ours & \textbf{0.874} & \textbf{0.829} & \textbf{0.897} & 0.855 & 0.843 & \textbf{0.891} & \textbf{0.797} & 0.865 & \textbf{0.857} & \textbf{0.888} & 0.871 & \textbf{0.943} & \textbf{0.821} & \textbf{0.864} \\
\bottomrule
\end{tabular}
}
\vspace{-2mm}
\caption{
\label{singleviewreconstructiontablelarge}
Single-view 3D reconstruction on ShapeNet. `Ours' is our full model \papertitle$_{Full}$. Our method outperforms other methods in the IoU (higher is better) and normal consistency (higher is better) metrics for most object categories, and performs competitively in the Chamfer-$L1$ (lower is better) metric. Taking the mean across all categories, \papertitle~outperforms all other methods in all three metrics.
}
\end{table*}

%% file: tables/multiviewshapenet.tex
\newcommand{\voxsign}{\scalerel*{\includegraphics{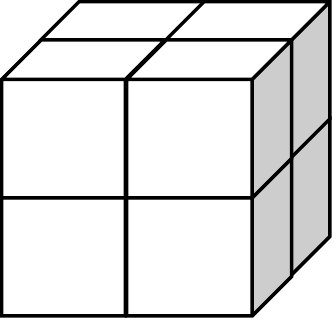}}{B}}

\newcommand{\voxcolor}{\rowfont{\color{gray}}}
\newcommand{\black}[1]{\textcolor{black}{#1}}

\begin{table*}[!ht]
\centering
\vspace{-1mm}
\resizebox{1.0\linewidth}{!}{%
\begin{tabu}
{c|c|ccccccccc}
\toprule
& Method & 1 view & 2 views & 3 views & 4 views & 5 views & 8 views & 12 views & 16 views & 24 views \\
\toprule
\voxcolor \multirow{5}{*}{\begin{tabular}{c}\black{(a) voxel measured} \\ \black{against voxel} \end{tabular}}
 & \black{3D-R2N2 \cite{choy20163d}} \voxsign & 0.560  & 0.603  & 0.617  & 0.625  & 0.634  & 0.635  & 0.636  & 0.636  & - \\
\voxcolor & \black{AttSets \cite{yang2020robust}} \voxsign  & 0.642  & 0.660  & 0.668  & 0.674  & 0.676  & 0.684  & 0.688  & 0.693 & 0.695 \\
\voxcolor & \black{Pix2Vox-A \cite{xie2019pix2vox}} \voxsign & 0.661  & 0.686  & 0.693  & 0.697  & 0.699  & 0.702  & 0.704  & 0.705  & 0.706 \\
\voxcolor & \black{Pix2Vox++/A \cite{xie2020pix2voxnew}} \voxsign & 0.670  & 0.695  & 0.704  & 0.708  & 0.711  & 0.715  & 0.717  & 0.718  & 0.719 \\
\voxcolor & \black{EVolT \cite{wang2021multi}} \voxsign & -  & -  & -  & 0.609  & -  & 0.698  & 0.720  & 0.729  & 0.738 \\
 \toprule
\multirow{9}{*}{\begin{tabular}{c}(b) measured\\ against mesh\end{tabular}}
 & Voxel upper-bound & \multicolumn{9}{c}{\textit{0.631}} \\
 & ONet$^\dagger$ \cite{mescheder2019occupancy} & 0.593  & -  & -  & -  & 0.621  & -  & -  & -  & - \\
 & 3D43D \cite{bautista2021generalization} & 0.621  & -  & -  & -  & 0.749  & -  & -  & -  & - \\
 & PiFU \cite{saito2019pifu} &  0.483  & 0.639  & 0.701 & 0.736  & 0.756  & 0.786  & 0.801  & 0.808  & 0.814 \\
 & \papertitle$_A$ & 0.634  & 0.706  & 0.733  & 0.749  & 0.759  & 0.775  & 0.784  & 0.789  & 0.793 \\
 & \papertitle$_B$ & 0.631 & 0.710 & 0.739 & 0.756 & 0.768 & 0.785 & 0.794 & 0.798 & 0.802 \\
 & \papertitle$_C$ & 0.660  & 0.740  & 0.766  & 0.778  & 0.786  & 0.799  & 0.805  & 0.808  & 0.810 \\
 & \papertitle$_D$ & 0.654  & 0.741  & 0.770  & 0.784  & 0.793  & 0.807  & 0.813  & 0.818  & 0.820 \\
 & \papertitle$_{Full}$ & \textbf{0.664}  & \textbf{0.750}  & \textbf{0.780}  & \textbf{0.795}  & \textbf{0.804}  & \textbf{0.819}  & \textbf{0.827}  & \textbf{0.831}  & \textbf{0.835} \\

\bottomrule
\end{tabu}%
}
\vspace{-2mm}
\caption{
\label{tab:multiviewtable}
Multi-view mean reconstruction IoU (higher is better) on ShapeNet. (a) presents the performance of the voxel-based methods, \textit{measured against the ground-truth voxels}.
All voxel based methods here generate $32^3$ resolution outputs. (b) presents the reconstruction performance \textit{measured against the actual ground truth meshes}.
First row of (b) shows the upper bound for voxel-based methods (using ground-truth voxels).
For any number of views, our methods vastly outperform all other methods in the mean IoU metric.
Our full model uses both 3D feature volume (similar to \papertitle$_D$), and pixel aligned local features, and achieves the best score in all measures.
\vspace{-5mm}
}
\end{table*}

%% file: tables/multiviewmodelnet.tex
\begin{table}[t!]
\centering
\vspace{-1mm}
\resizebox{\linewidth}{!}{
\begin{tabular}
{l@{\hspace{0.4em}}c@{\hspace{0.4em}}c@{\hspace{0.4em}}c@{\hspace{0.4em}}c@{\hspace{0.4em}}c@{\hspace{0.4em}}c@{\hspace{0.4em}}c}
\toprule
Method & 1 view & 2 views & 3 views & 5 views & 8 views & 12 views & 24 views \\
\midrule
PiFU \cite{saito2019pifu} &  0.530  & 0.637  & 0.690  & 0.736  & 0.769  & 0.783   & 0.799 \\
\papertitle$_{D}$ & 0.617  & 0.699  & 0.729 & 0.755  & 0.773 & 0.781  & 0.789 \\
\papertitle$_{Full}$ & \textbf{0.630}  & \textbf{0.709}  & \textbf{0.741}  & \textbf{0.769}  & \textbf{0.789}  & \textbf{0.798}  & \textbf{0.809} \\
\bottomrule
\end{tabular}%
}
\vspace{-2mm}
\caption{
\label{tab:multiviewmodelnettable}
Our method outperforms (IoU) PiFU on ModelNet reconstruction from any number of views.}
\vspace{-5mm}
\end{table}

%% file: conclusion.tex
\vspace{-1mm}
\section{Conclusion}
\vspace{-1mm}
We presented a unified single and multi-view 3D reconstruction framework \papertitle.
For 3D reconstruction, we use both: a 3D feature volume to capture the structure context, and pixel-aligned local features to provide high-detail local information.
We propose a 3D-structure-aware and multi-view aware architecture to fuse the 3D feature volumes across different images.
We show efficacy of interleaved 3D-UNet and multi-view transformer for this fusion.
We also propose to aggregate the multi-view fused pixel-aligned local features in a residual fashion, and demonstrate that the combined feature leads to the highest quality reconstruction.
Our experiments demonstrate the strength of our approach compared to existing approaches in shape reconstruction on multiple datasets.
Future works may include relaxation of correct camera pose requirement.
Finally, we anticipate that the proposed representation and fusion technique can be further extended to large scale scene reconstruction, depth fusion and novel view synthesis.

%% file: supplementary_materials/implicit_choice.tex
\section{Choice of Neural Implicit Shape Representation}
Shape representations like voxel, mesh and pointcloud represents the shape geometry explicitly.
In contrast, implicit shape representation approaches do not explicitly represent the geometry.
Instead, they typically represent some signal over the entire scene explicitly.
The shape geometry is thus implicitly encoded in the signal.

For example: one can represent occupancy as an implicit shape representation.
For any 3D point $X \in \mathbb{R}$, the occupancy function $f_{occ}(X) \in \{0, 1\}$ indicates whether $X$ is inside or outside the shape.
In such case, the shape surface is implicitly defined by the level set $0.5$:

\begin{equation}
    \left\{ X \in \mathbb{R}^3 : f_{occ}(X) = 0.5 \right\}
\end{equation}

Another option is to represent the signed distance function (SDF).
$f_{sdf}(X) \in \mathbb{R}$ specifies the distance of $X$ from the nearest shape surface.
$f_{sdf}(X)$ is also signed indicating if $X$ is inside or outside the shape.
The shape surface is implicitly defined by the level set $0$:

\begin{equation}
    \left\{ X \in \mathbb{R}^3 : f_{sdf}(X) = 0 \right\}
\end{equation}

Going from implicit to explicit representation is also easy.
For example, one can apply the Marching Cube algorithm \cite{lorensen1987marching,lewiner2003efficient} on the level set to obtain an explicit mesh representation.

Historically, these implicit functions $f_{occ}(.)$ and $f_{sdf}(.)$ have usually been represented in a discretized fashion in a grid.
More recently, neural implicit representation take a different approach, where a latent code or a grid of codes encode the shape.
A neural network based decoder input the latent shape code and a query 3D point $X$, and predicts the occupancy \cite{mescheder2019occupancy} or SDF \cite{park2019deepsdf} along that point.
Compared to the traditional discretized grid representation, this type of neural representation has the advantage in being able to represent the shape continuously.
Moreover, this type of representation is typically highly compact - allowing the storage of large amount of 3D data with reasonable space.
These representations are also highly suitable for neural network based manipulations for downstream task, while some explicit representations like polygonal mesh require complex manipulation operations for similar tasks.

In this work, I choose the neural implicit occupancy representation similar to ONet \cite{mescheder2019occupancy}.
However, unlike ONet that uses one latent code to encode a shape, I propose a different representation (discussed in Sec. 3 of the main paper) to improve the 3D shape representation and reconstruction.

%% file: tables/timingtable.tex
\begin{table}[!ht]
\centering
\resizebox{0.8\linewidth}{!}{
\begin{tabular}
{l|ccc}
\toprule
Method & 1 view & 8 views & 24 views \\
\midrule
ONet \cite{mescheder2019occupancy} & \textbf{0.61} & - & - \\
PiFU \cite{saito2019pifu} & 1.09 & 9.74 & 46.59 \\
\papertitle$_D$ & 0.74 & \textbf{0.80} & \textbf{1.03} \\
\papertitle$_{Full}$ & 5.47 & 21.21 & 54.31 \\
\bottomrule
\end{tabular}%
}
\caption{
\label{tab:timingtable}
Per object reconstruction time (seconds) using a GTX 1080Ti GPU. Lower is better.}
\end{table}

%% file: tables/singleviewpix3d.tex
\begin{table*}[!htb]
\centering
\resizebox{1.0\linewidth}{!}{
\begin{tabular}
{l|c@{\hspace{0.4em}}c@{\hspace{0.4em}}c@{\hspace{0.4em}}c|c@{\hspace{0.4em}}c@{\hspace{0.4em}}c@{\hspace{0.4em}}c|c@{\hspace{0.4em}}c@{\hspace{0.4em}}c@{\hspace{0.4em}}c}
\toprule
\multirow{2}{*}{\centering Category} & \multicolumn{4}{c}{IoU $\uparrow$} & \multicolumn{4}{c}{Chamfer-$L1 \downarrow$} & \multicolumn{4}{c}{Normal Consistency $\uparrow$} \\
                          & 3D-R2N2 & AtlasNet & ONet & Ours & 3D-R2N2 & AtlasNet & ONet & Ours & 3D-R2N2 & AtlasNet & ONet & Ours \\
\midrule
chair & 0.146 & - & 0.224 & \textbf{0.282} & 2.572 & \textbf{0.392} & 0.628 & 0.564 & 0.401 & \textbf{0.731} & 0.715 & 0.699 \\
desk & 0.170 & - & 0.268 & \textbf{0.342} & 1.776 & 0.509 & 1.047 & \textbf{0.418} & 0.478 & 0.717 & 0.682 & \textbf{0.764} \\
sofa & 0.483 & - & \textbf{0.580} & 0.557 & 0.537 & \textbf{0.367} & 0.427 & \textbf{0.367} & 0.645 & 0.792 & \textbf{0.817} & 0.800 \\
table & 0.137 & - & 0.204 & \textbf{0.288} & 2.665 & 0.556 & 1.071 & \textbf{0.484} & 0.431 & 0.768 & 0.720 & \textbf{0.772} \\
\midrule
mean & 0.234 & - & 0.319 & \textbf{0.367} & 1.888 & \textbf{0.456} & 0.793 & 0.458 & 0.489 & 0.752 & 0.733 & \textbf{0.759} \\

\bottomrule
\end{tabular}
}
\caption{
\label{singleviewreconstructionpix3d}
Single-view 3D reconstruction on Pix3D when trained only on ShapeNet. `Ours' refers to our model \papertitle$_{D}$, which outperforms or performs on par other methods.}
\end{table*}